\documentclass[10pt,twocolumn,letterpaper]{article}

\usepackage[pagenumbers]{cvpr} % To force page numbers, e.g. for an arXiv version

\definecolor{cvprblue}{rgb}{0.21,0.49,0.74}
\usepackage[pagebackref,breaklinks,colorlinks,allcolors=cvprblue]{hyperref}
\usepackage{multirow}
\usepackage[normalem]{ulem}
\useunder{\uline}{\ul}{}
\usepackage{arydshln}
\usepackage{booktabs}
\usepackage{colortbl}
\usepackage{xcolor}
\usepackage{graphicx}
\usepackage[accsupp]{axessibility}

\def\paperID{198} % *** Enter the Paper ID here
\def\confName{CVPR}
\def\confYear{2025}

\title{Latent Space Super-Resolution for Higher-Resolution \\ Image Generation with Diffusion Models}

\author{
Jinho Jeong, Sangmin Han, Jinwoo Kim, Seon Joo Kim\\
Yonsei University\\
{\tt\small \{3587jjh, seonjookim\}@yonsei.ac.kr}\\
}

\begin{document}

%------------------------------------------------------
\twocolumn[{%
\renewcommand\twocolumn[1][]{#1}%
\maketitle

\vspace{-1cm}
\begin{center}
    \centering
    \includegraphics[width=\textwidth]{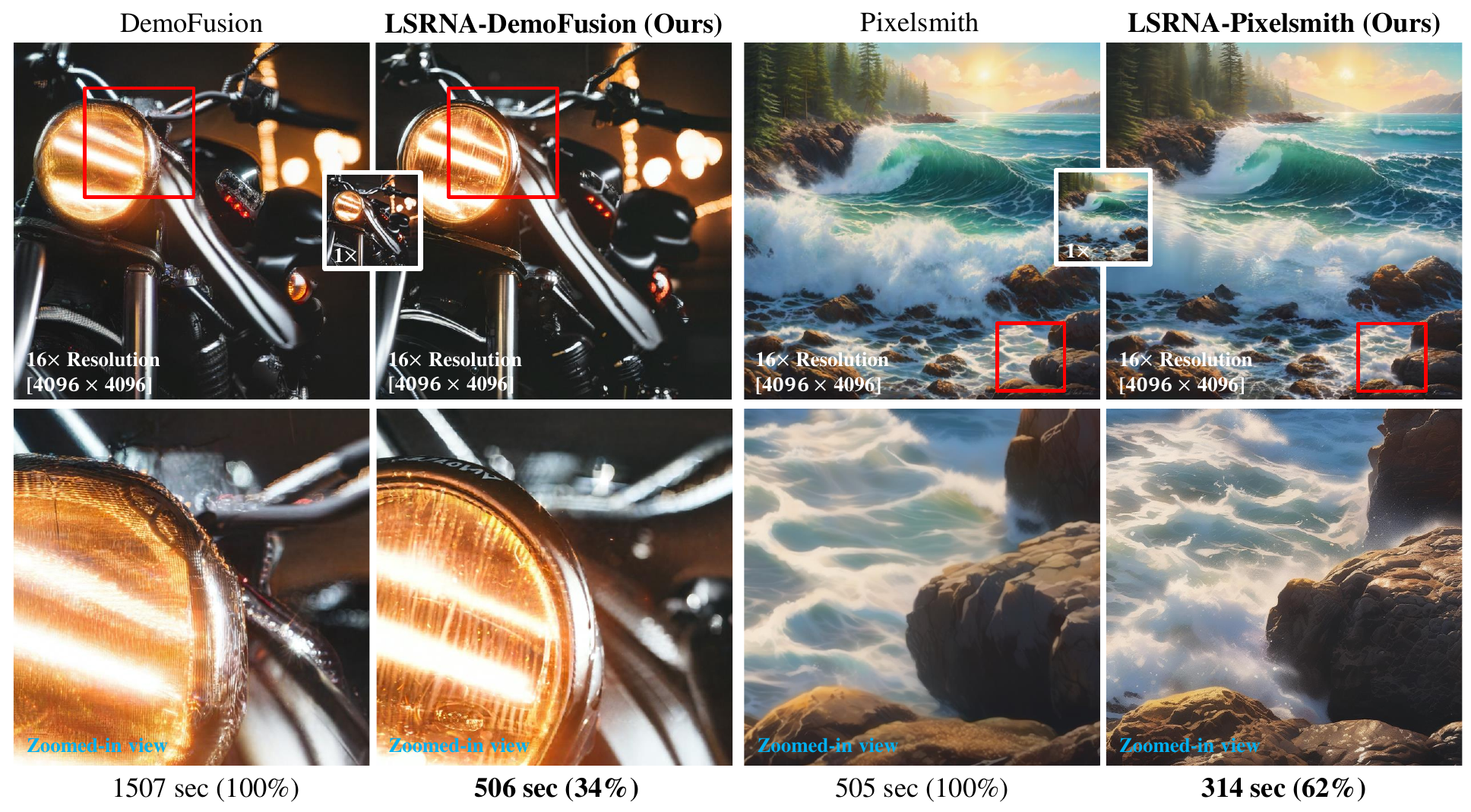}  
    \vspace{-1.5em}
    \captionof{figure}{\textbf{Comparisons of 16$\times$ image generation with and without LSRNA framework.} 
    Our proposed LSRNA framework improves reference-based higher-resolution image generation, enhancing detail and sharpness beyond the native resolution of SDXL~\cite{podell2023sdxl} ($1024^2$) while achieving faster generation speeds.}
    \label{fig:teaser}
\end{center}
}]
\begin{abstract} 
In this paper, we propose LSRNA, a novel framework for higher-resolution (exceeding 1K) image generation using diffusion models by leveraging super-resolution directly in the latent space.
Existing diffusion models struggle with scaling beyond their training resolutions, often leading to structural distortions or content repetition.
Reference-based methods address the issues by upsampling a low-resolution reference to guide higher-resolution generation. 
However, they face significant challenges: upsampling in latent space often causes manifold deviation, which degrades output quality. 
On the other hand, upsampling in RGB space tends to produce overly smoothed outputs.
To overcome these limitations, LSRNA combines Latent space Super-Resolution (LSR) for manifold alignment and Region-wise Noise Addition (RNA) to enhance high-frequency details.
Our extensive experiments demonstrate that integrating LSRNA outperforms state-of-the-art reference-based methods across various resolutions and metrics, while showing the critical role of latent space upsampling in preserving detail and sharpness.
The code is available at
\href{https://github.com/3587jjh/LSRNA}{https://github.com/3587jjh/LSRNA}.
\end{abstract}    
\section{Introduction}
\label{sec:intro}
\begin{figure}[t]
    \centering
    \includegraphics[width=\linewidth]{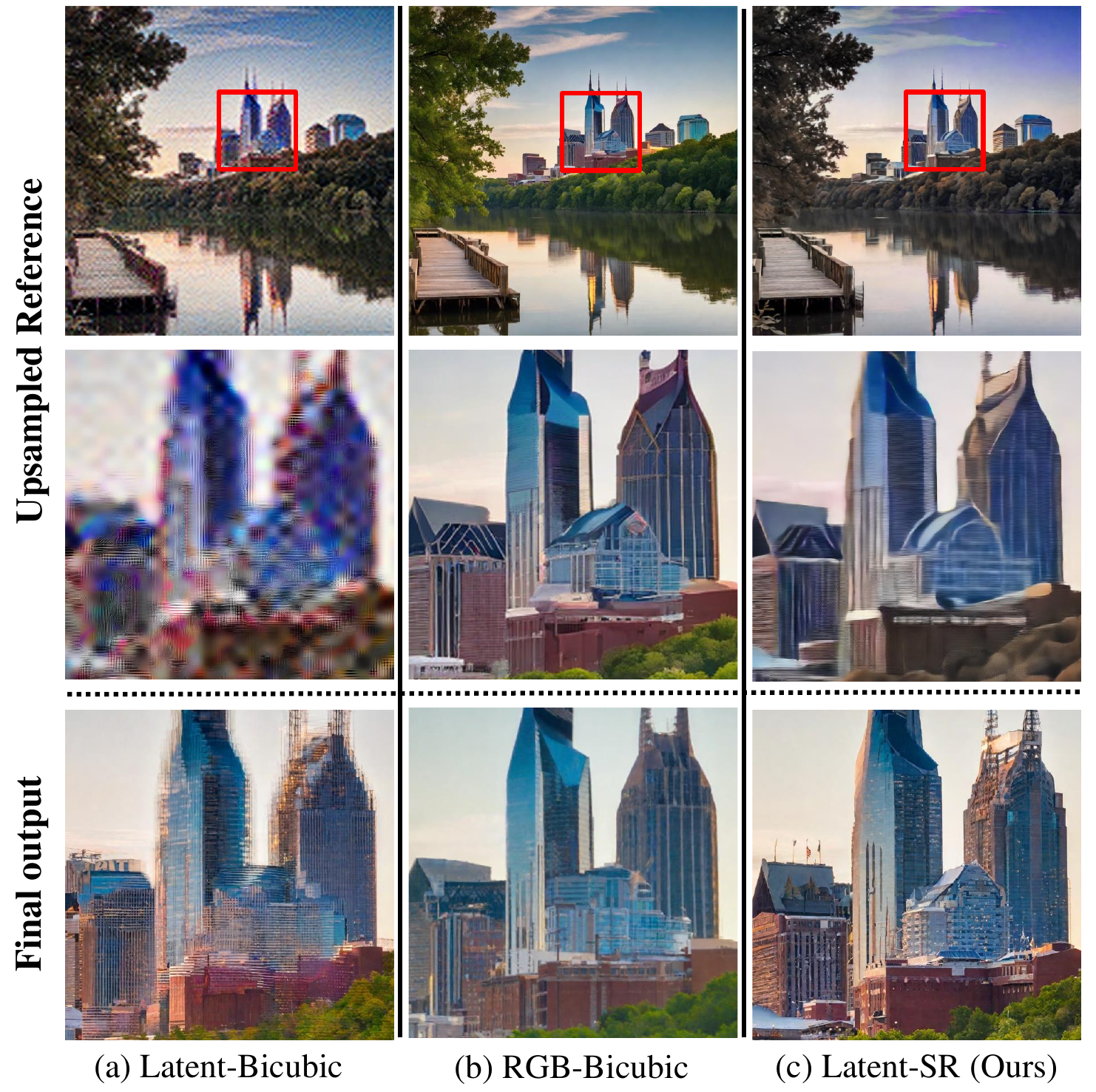}
    \caption{
        \textbf{Comparison of DemoFusion with different upsampling strategies.} 
        All methods are directly upsampled to 16$\times$ resolution.
        (a) Latent space bicubic upsampling causes manifold deviation, degrading output quality. 
        (b) RGB space bicubic upsampling produces outputs with reduced detail and sharpness. 
        (c) Our learned latent-space upsampling aligns the manifold, resulting in sharp and detailed outputs. 
        Best viewed \textbf{ZOOMED-IN.}
    }
    \label{fig:motivation}
\end{figure}

Diffusion models have rapidly become powerful tools in generative modeling, starting from foundational models like DDPM~\cite{ho2020denoising} and DDIM~\cite{song2020denoising}.
Latent Diffusion Models (LDMs)~\cite{rombach2022high} further revolutionized the field by reducing computational costs through operating in a lower-dimensional latent space.
These advancements have been applied to various tasks for image processing such as generation~\cite{nichol2021glide, saharia2022photorealistic, ramesh2022hierarchical}, editing~\cite{meng2021sdedit, hertz2022prompt, brooks2023instructpix2pix, kawar2023imagic}, and super-resolution~\cite{saharia2022image, ho2022cascaded}.

In parallel, with the rise of megapixel displays, demands for higher-resolution image generation (ranging from 2K to 8K) have grown drastically. 
However, existing diffusion models are limited to learning the distribution of fixed resolutions (\eg, $512^2$ for SD1.5), preventing them from generating images at unseen larger resolutions. 
Scaling diffusion models to larger resolutions presents several challenges. 
Training diffusion models from scratch~\cite{gu2023matryoshka, ding2023patched, hoogeboom2023simple, xie2024sana} or fine-tuning~\cite{zheng2024any, chen2024pixart, ren2024ultrapixel} for higher resolutions demands extensive data and computational resources.
Meanwhile, directly generating larger images using pretrained models leads to content repetition and structural distortions~\cite{bar2023multidiffusion, he2023scalecrafter}.
Alternatively, one can generate images at a pretrained resolution followed by super-resolution methods on RGB space. However, this approach tends to lack texture details and produce smoothed results~\cite{he2023scalecrafter, du2024demofusion}.

In response, recent studies have increasingly focused on training-free approaches that leverage pretrained diffusion models.
Among these, DemoFusion~\cite{du2024demofusion} stands out as a prominent method, laying the foundation for reference-based approaches~\cite{lin2024accdiffusion, shi2024resmaster, tragakis2024one}.
By generating a reference latent (or image) at the original size and upsampling it to guide the higher-resolution generation process, these approaches have demonstrated robust performance in producing high-quality images.
Nonetheless, what remains underexplored is the reference upsampling process, which makes it a key area for further investigation.

In this paper, we explore the significance of upsampled reference in reference-based image generation. 
Our first observation is that, as shown in \Cref{fig:motivation}(a), the bicubic-upsampled reference latent appears visually unappealing, exhibiting artifacts that suggest a deviation from the target resolution's manifold. 
This deviation poses a considerable problem as it interferes with the subsequent generation process by providing suboptimal guidance, preventing the output from properly aligning with the desired manifold. 
Existing reference-based methods with latent space upsampling~\cite{du2024demofusion, lin2024accdiffusion} mitigate the issue by gradually increasing image resolution in stages (\ie, progressive upscaling). 
However, it leads to slower inference times and the potential accumulation of errors in the content.

Given these limitations, we explored RGB space upsampling as an alternative, inspired by reference-based RGB upsampling methods~\cite{shi2024resmaster, tragakis2024one}. 
This approach has the potential advantage of avoiding the artifacts observed in latent space upsampling, as RGB-based methods work directly in pixel space, where artifacts may be less pronounced.
Specifically, we replace latent upsampling with RGB upsampling, where a reference latent is decoded into an RGB image, bicubic-upsampled, and then re-encoded back into the latent space. 
However, \Cref{fig:motivation}(b) shows that the RGB space upsampling produces smooth and less detailed outputs compared to the original latent upsampling.

Based on these observations, we hypothesize that upsampling within the latent space plays a crucial role in preserving sharpness and detail essential for high-resolution image generation. However, the issue of manifold deviation caused by upsampling remains insufficiently addressed.
To tackle this, we propose a novel approach: learning to map low-resolution latent representations onto the higher-resolution manifold.
We introduce a lightweight Latent space Super-Resolution (LSR) module that operates directly within the latent space. 
Additionally, to facilitate the generation of high-frequency details in the subsequent generation process, we propose the Region-wise Noise Addition (RNA) module. 
Inspired by SDEdit~\cite{meng2021sdedit}, which improves image quality through uniform noise injection followed by denoising, RNA adaptively injects Gaussian noise into specific regions of the upsampled reference latent. 

\begin{figure}[t]
    \centering
    \includegraphics[width=\linewidth]{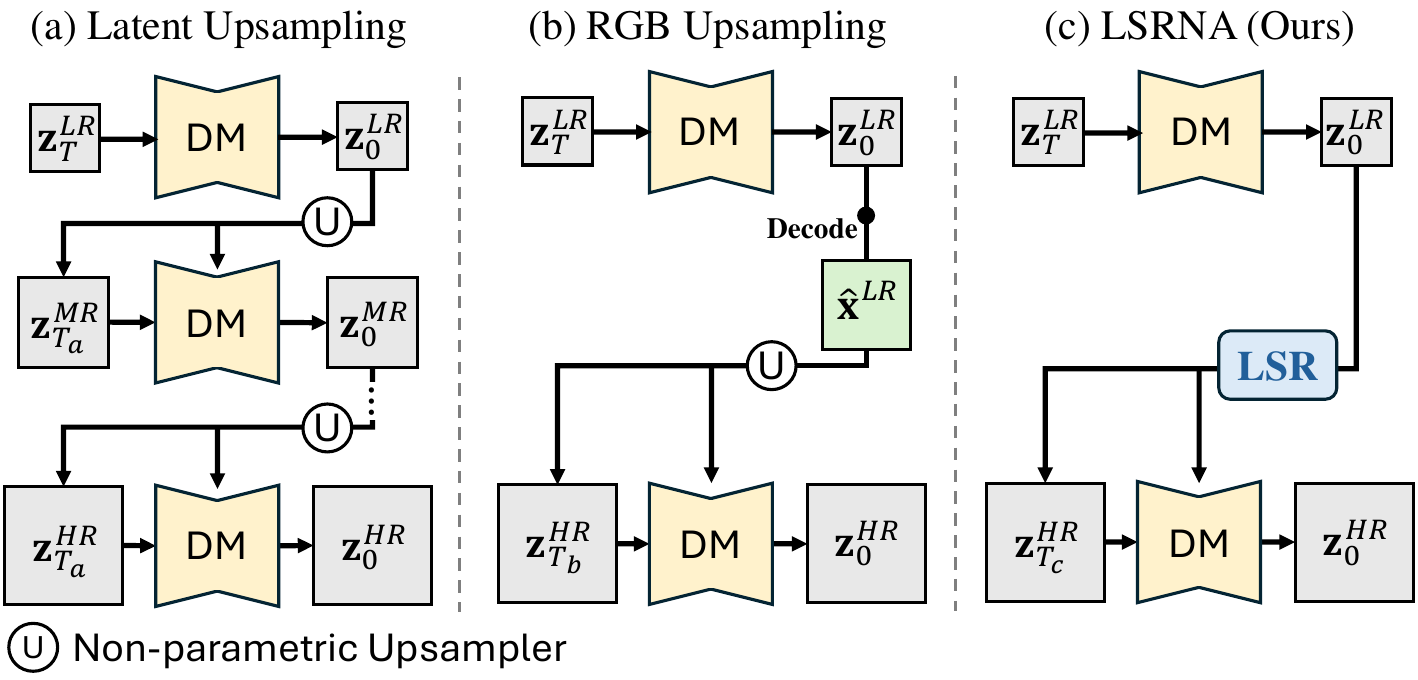}
    \vspace{-1em}
    \caption{
    \textbf{Framework Comparison.} 
    (a) Existing latent upsampling framework rely on progressive upsampling to address manifold deviation.  
    (b) Existing RGB upsampling framework can directly upsample (optionally progressively), but produce smooth output.
    (c) Our framework enables latent upsampling without progressive upscaling with much fewer denoising steps ($T_c<T$) while producing detailed outputs (RNA omitted for simplicity).
    LR, MR, HR: low/mid/high resolution; DM: Diffusion Model.
    }
    \label{fig:framework_comparison}
\end{figure}
Combining the manifold alignment capability of LSR with the detail generation ability of RNA, our framework, \textbf{LSRNA}, is the first latent upsampling framework to eliminate the need for progressive upscaling as illustrated in \Cref{fig:framework_comparison}.
Our high-quality guidance also allows for fewer subsequent denoising steps, enabling faster generation with minimal quality loss.
We validate the effectiveness of LSRNA by incorporating it into the state-of-the-art latent upsampling-based method~\cite{du2024demofusion}, pushing the boundaries of high-resolution image generation. 
Furthermore, we demonstrate that latent upsampling is crucial for preserving sharpness and detail by applying LSRNA to a very recent RGB upsampling-based model~\cite{tragakis2024one}.
\section{Related Works}
\noindent\textbf{Latent Diffusion Models.}
Diffusion models, with foundations in DDPM~\cite{ho2020denoising}, DDIM~\cite{song2020denoising}, and ADM~\cite{dhariwal2021diffusion}, have become a dominant approach for image generation. 
Latent Diffusion Models (LDMs)~\cite{rombach2022high} further advance the field by shifting the diffusion process into the latent space, using a pretrained autoencoder to compress images.
This approach allows for faster computations and lower memory usage, supporting a wide range of applications~\cite{hertz2022prompt, gu2022vector, saharia2022palette, ruiz2023dreambooth, brooks2023instructpix2pix, zhang2023adding, mokady2023null, podell2023sdxl}. 
Notably, SD~\cite{rombach2022high} and SDXL~\cite{podell2023sdxl}, both types of LDMs, have become widely adopted for their robustness and efficiency in generating images from text prompts. 
SD and SDXL are widely used for high-resolution image generation, laying the basis for the methods discussed in the following sections.

\vspace{1mm}
\noindent\textbf{Higher-resolution Image Generation.}
While LDMs produce impressive results at their pretrained resolution, attempting to generate images beyond this scale often leads to repetitive patterns and structural inconsistencies, significantly lowering image quality. 
Various studies have explored ways for diffusion models to adapt to higher-resolution images, either through training from scratch~\cite{gu2023matryoshka, ding2023patched, hoogeboom2023simple, xie2024sana} or fine-tuning~\cite{zheng2024any, ren2024ultrapixel, chen2024pixart} on larger-resolution datasets. However, these approaches still require substantial data collection and computational resources.
%; \eg, training PixArt-$\Sigma$~\cite{chen2024pixart} at a 4K resolution needs 16 A100 GPUs.

Consequently, numerous approaches ~\cite{bar2023multidiffusion, he2023scalecrafter, haji2024elasticdiffusion, huang2024fouriscale, du2024demofusion, zhang2023hidiffusion, lin2024accdiffusion, tragakis2024one, shi2024resmaster} have leveraged pretrained models as alternatives without requiring additional training. 
ScaleCrafter~\cite{he2023scalecrafter}, HiDiffusion~\cite{zhang2023hidiffusion}, and FouriScale~\cite{huang2024fouriscale} use dilated convolutions in certain U-Net layers to tackle object repetition.
%, but denoising high-resolution latents in a single pass requires considerable GPU memory.
Meanwhile, MultiDiffusion~\cite{bar2023multidiffusion} independently denoises patches of the original resolution and aggregates them to produce a panoramic image. 
%Although it maintains consistent memory usage, it lacks global structural constraints, leading to persistent object repetition.
ElasticDiffusion~\cite{haji2024elasticdiffusion} revisits classifier-free guidance, enhancing global coherence by separating global and local terms.
DemoFusion~\cite{du2024demofusion} has emerged as a prominent method for its reference-based approach, generating a reference latent at the original resolution and upsampling it to guide the higher-resolution generation process, inspiring later studies~\cite{lin2024accdiffusion, shi2024resmaster, tragakis2024one}. %Among these, Pixelsmith ~\cite{tragakis2024one} successfully generates gigapixel-scale images with slider-based guidance and masking but shows reduced sharpness compared to previous methods.

While training-free methods effectively improve higher-resolution image generation, the absence of model adaptation to higher resolutions prior to inference can result in suboptimal image quality.
Self-Cascade~\cite{guo2024make} enables fast model adaptation with minimal parameters by utilizing learnable, time-aware feature upsampler modules to incorporate knowledge from a small set of newly acquired high-quality images.
Our method achieves even faster adaptation, driven by a different motivation: mapping low-resolution latent onto the higher-resolution manifold.
\section{Preliminary}
\noindent\textbf{Latent Diffusion Model.}
In Latent Diffusion Model (LDM), an autoencoder is composed of an encoder \(\mathcal{E}\) and a decoder \(\mathcal{D}\).
Given an image \(\mathbf{x}\), the encoder \(\mathcal{E}\) maps \(\mathbf{x}\) to the latent space as \(\mathbf{z} = \mathcal{E}(\mathbf{x})\).
The LDM then operates within this latent space, utilizing two processes, a forward process and a denoising process.
In the forward process, Gaussian noise is incrementally added to \(\mathbf{z}_0 (= \mathbf{z}\)) across a fixed number of timesteps \(T\), transforming it to approximate a Gaussian distribution \(\mathcal{N}(0, I)\) as:
\begin{equation}
q(\mathbf{z}_t|\mathbf{z}_{t-1}) = \mathcal{N}(\mathbf{z}_t; \sqrt{1 - \beta_t} \mathbf{z}_{t-1}, \beta_t \mathbf{I}),
\end{equation}
where \(\{\beta_t\}_{t=1...T}\) is a predefined variance schedule.
The denoising process begins with \(\mathbf{z}_T\) and gradually removes the added noise to recover the original latent \(\mathbf{z}_0\). This is achieved through a learnable noise-prediction network~\cite{dhariwal2021diffusion, dosovitskiy2020image} parameterized by \(\theta\), described as:
\begin{equation}
p_\theta(\mathbf{z}_{t-1}|\mathbf{z}_t) = \mathcal{N}(\mathbf{z}_{t-1}; \mathbf{\mu}_\theta(\mathbf{z}_t, t), \mathbf{\Sigma}_\theta(\mathbf{z}_t, t)).
\end{equation}
After denoising, the decoder \(\mathcal{D}\) transforms the restored latent back into an image \(\hat{\mathbf{x}}\).

\vspace{1mm}
\noindent\textbf{Reference-based Image Generation.}
In reference-based approaches, an original-size reference is upsampled to guide the higher-resolution generation process. 
Additionally, this upsampling-then-generation process can be divided into smaller incremental steps, progressively increasing the resolution to enhance precision.
For convenience, let LR denote the resolution of the current stage and HR denote the resolution of the next stage, considering the progressive upsampling procedure.

Existing reference-based methods belongs to either the latent upsampling framework or the RGB upsampling framework.
To guide the subsequent generation process, the former employs the upsampled latent $\mathbf{z'}_0^{HR} = U_{np}(\mathbf{z}_0^{LR})$ as guidance, where $U_{np}$ represents a non-parametric upsampler (\eg, bicubic interpolation).
In contrast, the latter first decodes $\mathbf{z}_0^{LR}$ into an image $\hat{\mathbf{x}}^{LR}$, which is then upsampled to $\hat{\mathbf{x}}^{HR}$ and used as guidance.

The denoising process for higher-resolution generation begins by injecting Gaussian noise into the upsampled guidance to initialize \(\mathbf{z}_{T_{init}}^{HR}\), where \(T_{init} \in [1, T]\) is the starting timestep.
At each timestep \(t \in [1, T_{init}]\), the denoised latent \(\mathbf{z}_t^{HR}\) is updated with guidance, defined by:
\begin{equation}
\mathbf{z}_t^{HR} \leftarrow \Phi_t(\mathbf{z}_t^{HR}, \mathbf{g}^{HR}),
\end{equation}
where $\mathbf{g}^{HR}$ represents the guidance ($\mathbf{z'}_0^{HR}$ or $\hat{\mathbf{x}}^{HR}$) specific to each framework, and \(\Phi_t\) is an updating function that integrates the reference information.
The specific denoising strategy for handling the larger latent size varies across reference-based methods (\eg, dividing it into overlapping patches~\cite{du2024demofusion}).
\section{Methodology}
\begin{figure}[t]
    \centering
    \includegraphics[width=\linewidth]{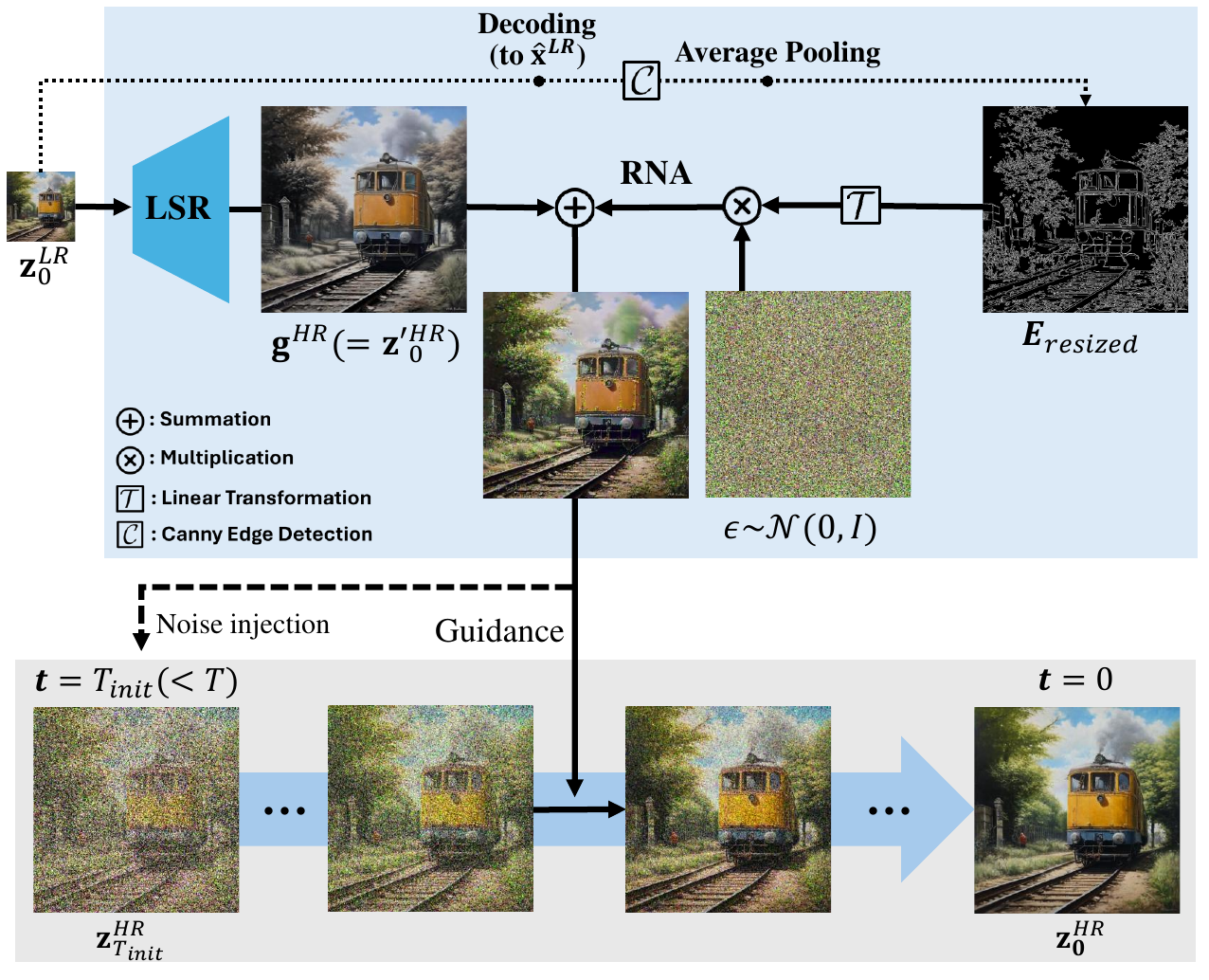}
    \vspace{-1em}
    \caption{
    \textbf{Overview of LSRNA.}
    The proposed LSRNA enhances reference upsampling with Latent space Super-Resolution (LSR) and Region-wise Noise Addition (RNA). 
    LSR directly maps the low-resolution reference latent onto the high-resolution manifold. 
    RNA then injects region-adaptive noise into the mapped reference, guided by a canny edge map.
    RNA facilitates detail generation in the higher-resolution generation stage.
    }
    \label{fig:overview}
\end{figure}
An overview of LSRNA framework is illustrated in \Cref{fig:overview}.
LSRNA consists of Latent space Super-Resolution (LSR) module and Region-wise Noise Addition (RNA) module, providing better guidance.
Given $\mathbf{z}_{0}^{LR}$ (\ie, low-resolution reference latent), LSR maps the $\mathbf{z}_{0}^{LR}$ into a higher-resolution manifold, resulting in $\mathbf{g}^{HR}(=\mathbf{z'}_{0}^{HR})$.
Then, RNA constructs region-adaptive Gaussian noise, which is added to the upsampled reference $\mathbf{g}^{HR}$, facilitating detail generation during the subsequent high-resolution generation process.
The following sections provide a detailed explanation of each module.

\subsection{Latent Space Super-Resolution}
\noindent\textbf{Architecture.}
The LSR module consists of two main components: a backbone and an upsampler.
The backbone takes the LR latent $\mathbf{z}_0^{LR} \in \mathbb{R}^{h \times w \times C}$ as input and extracts a feature map $F \in \mathbb{R}^{h \times w \times C'}$.
Here, any lightweight SR network~\cite{zhang2018image, liang2021swinir} can be used as backbone, modifying its input and output channels to the latent channel dimension $C$. 

For the upsampler, we employ LIIF~\cite{chen2021learning}, an MLP-based upsampler that enables flexible upsampling to arbitrary resolutions with a single training.
For a pixel $q$ in the target HR latent $\mathbf{g}^{HR} \in \mathbb{R}^{H \times W \times C}$, let its spatial coordinate denoted as $q_{coord} \in \mathbb{R}^2$. 
The upsampler $U$ takes the feature $F$ and $q_{coord}$ as input and predicts latent value for the pixel $q$ as:
\begin{equation}
\mathbf{g}^{HR}(q) = U(F, q_{coord} ; \theta_U),
\end{equation}
where \(\theta_U\) denotes the parameters of the upsampler.
The entire $\mathbf{g}^{HR}$ can be predicted by querying all pixels in $\mathbf{g}^{HR}$.

\vspace{1mm}
\noindent\textbf{Training Setup.}
The LSR module takes an LR latent as input and outputs an HR latent. 
Ideally, obtaining LR-HR latent training pairs directly from the diffusion process would be advantageous. 
However, acquiring a ground-truth HR latent that both preserves the same content as the LR latent and aligns with the HR manifold is infeasible.

To address this, we use the real-world dataset~\cite{kuznetsova2020open} to obtain ground-truth HR RGB images.
Specifically, we construct training pairs through two steps: (i) applying  degradation (\ie, downsampling) to the HR images to generate LR RGB images, and (ii) separately encoding both the HR and LR RGB images using a pretrained encoder $\mathcal{E}$.
The design of steps (i) and (ii) mainly focuses on reducing the domain gap between training and inference, which is discussed in the appendix.
To support multi-scale training, we introduce variability by randomly cropping HR images to various sizes and downscaling them by multiple factors during step (i).
Our data preparation strategy produces a training dataset with a total of 4.7M LR-HR latent pairs.
Using this setup, the LSR module is trained with $L_1$ loss.

\subsection{Region-wise Noise Addition}
In addition to LSR, we propose Region-wise Noise Addition (RNA) to further enhance finer details. 
The motivation for RNA arises from the observation that certain high-frequency regions in output images still have room for improvement.
This may stem from the use of $L_1$ training loss for the LSR, which tends to favor pixel-wise averaged solutions and often results in smoothed predictions~\cite{bruna2015super, dosovitskiy2016generating, johnson2016perceptual}.
We hypothesize that introducing controlled variations in the latent values according to region-wise property (\eg, frequency) can provide better guidance for generating finer details. 
To achieve this, we design RNA to adaptively add Gaussian noise to specific areas of the upsampled reference latent, focusing on detail-critical regions.

We adopt Canny edge detection~\cite{canny1986computational} to identify high-frequency regions. 
It is preferred over other edge detection methods~\cite{gabor1946theory, marr1980theory} due to its ability to capture both weak and strong edges, as further discussed in the appendix.
To obtain the edge map, we first decode the reference latent $\mathbf{z}_0^{LR} \in \mathbb{R}^{h \times w \times C}$ into an image \(\hat{\mathbf{x}}^{LR} \in \mathbb{R}^{sh \times sw \times 3}\), where $s$ denotes the compression ratio of the autoencoder.
Next, we apply Canny edge detection to \(\hat{\mathbf{x}}^{LR}\) to obtain the edge map $\mathbf{E}$. 
The edge map is then resized to match the target latent size using an average pooling, resulting in $\mathbf{E}_{resized} \in \mathbb{R}^{H \times W}$.

To flexibly adjust Gaussian noise, we normalize the edge map values using a linear transformation $\mathcal{T}$ to map the edge values to a predefined range $[e_{min}, e_{max}]$.
The normalized edge map is then used to modulate Gaussian noise $\epsilon \sim \mathcal{N}(0,I)$.
The resulting region-wise noise is added to the latent immediately after it is upsampled by the LSR module. 
This process can be expressed as:
\begin{equation}
\mathbf{g}^{HR} \leftarrow \mathbf{g}^{HR} + \mathcal{T}(\mathbf{E}_{resized}) \cdot \epsilon,
\end{equation}
where $\mathbf{g}^{HR}$ is the upsampled latent from the LSR module.
\section{Experiments}
\begin{table*}[]
\caption{
\textbf{Quantitative comparison results on OpenImages-Test.} 
The best and second-best performances are highlighted in \textcolor[HTML]{FF0000}{red} and \textcolor[HTML]{0000FF}{blue}, respectively. 
Methods above the dashed line are non-reference-based, while those below are reference-based.
FouriScale is not measured above 2K due to out-of-memory on our V100 GPU.
}
\vspace{-1mm}
\resizebox{\textwidth}{!}{%%
\renewcommand{\arraystretch}{1.1}
\begin{tabular}{c|l|cccccc|c}
\toprule
Resolution                   & Method             & FID (↓)                      & KID (↓)                       & pFID (↓)                     & pKID (↓)                      & IS (↑)                       & CLIP (↑)                     & Time (sec)                 \\ \bottomrule \toprule
                             & SDXL$+$BSRGAN \cite{zhang2021designing}      & 84.32                        & 0.0080                        & 39.71                        & 0.0090                        & 30.11                        & 0.303                        & {\color[HTML]{FF0000} 16}  \\
                             & SDXL (Direct) \cite{podell2023sdxl}      & 113.19                       & 0.0222                        & 64.20                        & 0.0198                        & 19.13                        & 0.294                        & 80                         \\
                             & ScaleCrafter \cite{he2023scalecrafter}       & 91.76                        & 0.0103                        & 45.60                        & 0.0106                        & 29.09                        & 0.301                        & 101                        \\
                             & FouriScale \cite{huang2024fouriscale}        & 104.30                       & 0.0175                        & 57.00                        & 0.0167                        & 23.90                        & 0.300                        & 134                        \\
                             & HiDiffusion \cite{zhang2023hidiffusion}       & 90.23                        & 0.0106                        & 43.76                        & 0.0099                        & 27.17                        & 0.299                        & {\color[HTML]{0000FF} 50}  \\ \cdashline{2-9} 
                             & Self-Cascade \cite{guo2024make}      & {\color[HTML]{FF0000} 83.50} & {\color[HTML]{FF0000} 0.0064} & {\color[HTML]{FF0000} 36.44} & {\color[HTML]{0000FF} 0.0070} & 31.56                        & {\color[HTML]{FF0000} 0.305} & 90                         \\ 
                             & DemoFusion \cite{du2024demofusion}        & 85.02                        & 0.0079                        & 38.96                        & 0.0087                        & {\color[HTML]{FF0000} 32.54} & 0.302                        & 205                        \\
                             & \textbf{LSRNA-DemoFusion (Ours)} & {\color[HTML]{0000FF} 83.58} & {\color[HTML]{0000FF} 0.0077} & {\color[HTML]{0000FF} 36.55} & {\color[HTML]{FF0000} 0.0069} & {\color[HTML]{0000FF} 31.74} & 0.303 & 115                        \\  
                             & Pixelsmith \cite{tragakis2024one}         & 85.40 & 0.0091 & 39.51 & 0.0087 & 30.84 & {\color[HTML]{0000FF} 0.304} & 126                        \\
\multirow{-10}{*}{2048$\times$2048} & \textbf{LSRNA-Pixelsmith (Ours)} & 83.90                        & 0.0079                        & 37.33                        & 0.0074                        & 30.75                        & 0.302                        & 86                         \\ \bottomrule \toprule
                             & SDXL$+$BSRGAN \cite{zhang2021designing}     & 80.71                        & 0.0058                        & 39.81                        & 0.0122                        & 25.95                        & 0.294                        & {\color[HTML]{FF0000} 17}  \\
                             & SDXL (Direct) \cite{podell2023sdxl}      & 134.57                       & 0.0370                        & 71.73                        & 0.0230                        & 12.57                        & 0.276                        & 245                        \\
                             & ScaleCrafter \cite{he2023scalecrafter}       & 93.32                        & 0.0142                        & 43.35                        & 0.0130                        & 23.12                        & 0.288                        & 419                        \\
                             & HiDiffusion \cite{zhang2023hidiffusion}        & 101.10                       & 0.0177                        & 51.66                        & 0.0158                        & 21.90                        & 0.283                        & {\color[HTML]{0000FF} 129} \\ \cdashline{2-9} 
                             & Self-Cascade \cite{guo2024make}       & {\color[HTML]{FF0000} 78.34} & {\color[HTML]{FF0000} 0.0042} & 34.40                        & 0.0076                        & 25.93                        & 0.294                        & 210                        \\ 
                             & DemoFusion \cite{du2024demofusion}        & 83.63                        & 0.0072                        & 38.64                        & 0.0082                        & {\color[HTML]{0000FF} 27.74} & 0.294                        & 648                        \\
                             & \textbf{LSRNA-DemoFusion (Ours)} & 80.53 & 0.0057 & {\color[HTML]{0000FF} 33.31} & {\color[HTML]{0000FF} 0.0064} & 27.17 & {\color[HTML]{FF0000} 0.297} & 223                        \\ 
                             & Pixelsmith \cite{tragakis2024one}       & 81.21                        & 0.0069                        & 40.86 & 0.0108 & 25.59                        & 0.294 & 232                        \\
\multirow{-9}{*}{2048$\times$4096}  & \textbf{LSRNA-Pixelsmith (Ours)} & {\color[HTML]{0000FF} 78.47} & {\color[HTML]{0000FF} 0.0050} & {\color[HTML]{FF0000} 32.96} & {\color[HTML]{FF0000} 0.0062} & {\color[HTML]{FF0000} 28.04} & {\color[HTML]{0000FF} 0.295} & 157                        \\ \bottomrule \toprule
                             & SDXL$+$BSRGAN \cite{zhang2021designing}      & {\color[HTML]{0000FF} 84.64} & 0.0081                        & 37.04                        & 0.0149                        & 30.13                        & 0.302                        & {\color[HTML]{FF0000} 17}  \\
                             & SDXL (Direct) \cite{podell2023sdxl}     & 217.88                       & 0.0976                        & 99.05                        & 0.0468                        & 9.15                         & 0.270                        & 786                        \\
                             & ScaleCrafter \cite{he2023scalecrafter}       & 110.49                       & 0.0202                        & 54.91                        & 0.0196                        & 21.80                        & 0.293                        & 1351                       \\
                             & HiDiffusion \cite{zhang2023hidiffusion}        & 128.28                       & 0.0319                        & 100.85                       & 0.0564                        & 19.62                        & 0.280                        & {\color[HTML]{0000FF} 240} \\ \cdashline{2-9} 
                             & Self-Cascade \cite{guo2024make}       & 90.94                        & 0.0106                        & 43.91                        & 0.0187                        & 27.22                        & 0.300                        & 669                        \\  
                             & DemoFusion \cite{du2024demofusion}        & 87.29                        & 0.0089                        & 32.89                        & 0.0102                        & 29.69                        & 0.300                        & 1507                       \\
                             & \textbf{LSRNA-DemoFusion (Ours)} & 85.03 & {\color[HTML]{0000FF} 0.0077} & {\color[HTML]{FF0000} 29.12} & {\color[HTML]{FF0000} 0.0085} & {\color[HTML]{0000FF} 31.50} & {\color[HTML]{FF0000} 0.304} & 506                        \\ 
                             & Pixelsmith \cite{tragakis2024one}        & 84.75                        & 0.0086 & 32.34 & 0.0111 & 30.21 & {\color[HTML]{0000FF} 0.303} & 505                        \\
\multirow{-9}{*}{4096$\times$4096}  & \textbf{LSRNA-Pixelsmith (Ours)} & {\color[HTML]{FF0000} 84.19} & {\color[HTML]{FF0000} 0.0075} & {\color[HTML]{0000FF} 29.62} & {\color[HTML]{0000FF} 0.0090} & {\color[HTML]{FF0000} 31.74} & 0.302                        & 313                        \\ \bottomrule
\end{tabular}
}%%
\vspace{-4mm}
\label{tab:main}
\end{table*}
In this section, we present both quantitative and qualitative results, followed by ablation studies.
Additional experiments and results, including the importance of latent space upsampling and
the training details of the LSR module can be found in the appendix.

\subsection{Experiment Settings}
We adopt several existing higher-resolution image generation methods, categorized by whether they use a reference.
Non-reference-based methods include SDXL$+$BSRGAN~\cite{zhang2021designing}, which generates a 1K resolution image using SDXL and upscales it with BSRGAN, and SDXL (Direct)~\cite{podell2023sdxl}, which performs direct inference at the target resolution. 
Methods such as ScaleCrafter~\cite{he2023scalecrafter}, FouriScale~\cite{huang2024fouriscale}, and HiDiffusion~\cite{zhang2023hidiffusion} are also included, all of which utilize dilated convolutions for denoising. 
Reference-based methods include DemoFusion~\cite{du2024demofusion}, the foundational reference-based approach, Pixelsmith~\cite{tragakis2024one}, which use a slider mechanism for structural coherence, and Self-Cascade~\cite{guo2024make}, which utilizes learnable time-aware upsample modules.
To evaluate the effectiveness of our LSRNA framework, we integrate existing reference-based methods into our framework, resulting in two configurations: \textbf{LSRNA-DemoFusion} and \textbf{LSRNA-Pixelsmith}. 
Each configuration uses the denoising strategy of its respective model during the higher-resolution generation phase.

All methods are based on the SDXL, with inference and runtime measurements conducted on a single NVIDIA Tesla V100-SXM2 GPU.
We set the denoising steps in the LSRNA framework to 30 DDIM steps, while existing methods use the standard 50, which is further discussed in the ablation study.

\subsection{Quantitative Results}
\noindent\textbf{Metrics.}
We employ four widely-used metrics: FID~\cite{heusel2017gans}, KID~\cite{binkowski2018demystifying}, IS~\cite{salimans2016improved}, and CLIP Score~\cite{radford2021learning}, to evaluate the quality and semantic alignment of generated images. FID and KID require resizing images to $299^2$ to match the input size of the Inception network~\cite{szegedy2015going}. 
However, the resizing can result in a loss of high-resolution details, potentially affecting the evaluation of high-resolution images. 
To address this limitation, inspired by Anyres-GAN~\cite{chai2022anyresolution}, we also compute FID and KID on 50,000 randomly cropped 1K-resolution patches from both real and generated images. 
We term these patch-based metrics pFID and pKID, as they better capture the finer details present in high-resolution images.

\vspace{1mm}
\noindent\textbf{Dataset Preparation.}
We sample a validation set of 400 images and a test set of 1,000 images from the OpenImages dataset~\cite{kuznetsova2020open}, naming them OpenImages-Valid and OpenImages-Test. 
Captions for both the validation and test sets are generated using BLIP2~\cite{li2023blip}. 
Since finer details in the generated patches may be inadvertently penalized when reference patches are derived from low-resolution ground-truth images, we filter both sets to include only high-resolution images (larger than 3K) to ensure reliable evaluation.
We also note that although the training dataset for our LSR module is also sampled from OpenImages, we ensure that the training, validation, and test sets are completely independent, with no overlap in image IDs.

\vspace{1mm}
\noindent\textbf{Comparison.}
As shown in \Cref{tab:main}, the proposed LSRNA framework meaningfully enhances the performance of both DemoFusion and Pixelsmith across resolutions and metrics.
For DemoFusion, a latent upsampling-based method, integrating LSRNA eliminates the need for progressive upscaling and reduces denoising steps by providing high-quality guidance.
These changes result in faster inference times and performance improvements in most cases, demonstrating LSRNA’s capability to optimize existing methods effectively. 
Furthermore, Pixelsmith, an RGB upsampling-based method, also benefits from LSRNA integration. 
By shifting to latent upsampling-based guidance, Pixelsmith achieves improved performance across resolutions and metrics, while the reduced denoising depth introduced by LSRNA accelerates inference.

Compared to non-reference-based methods, reference-based methods achieve better performance, but at the cost of slower inference times. 
Non-reference-based methods, \eg, SDXL$+$BSRGAN and HiDiffusion, offer faster generation speeds, however, struggle to maintain semantic alignment and perceptual fidelity. 
Meanwhile, LSRNA-DemoFusion and LSRNA-Pixelsmith achieve state-of-the-art performance across various metrics while maintaining fast generation times. 
These results highlight the efficacy and adaptability of the LSRNA framework.
%, pushing boundaries of high-resolution image generation.

\subsection{Qualitative Results}
\begin{figure*}[t]
    \centering
    \includegraphics[width=\linewidth]{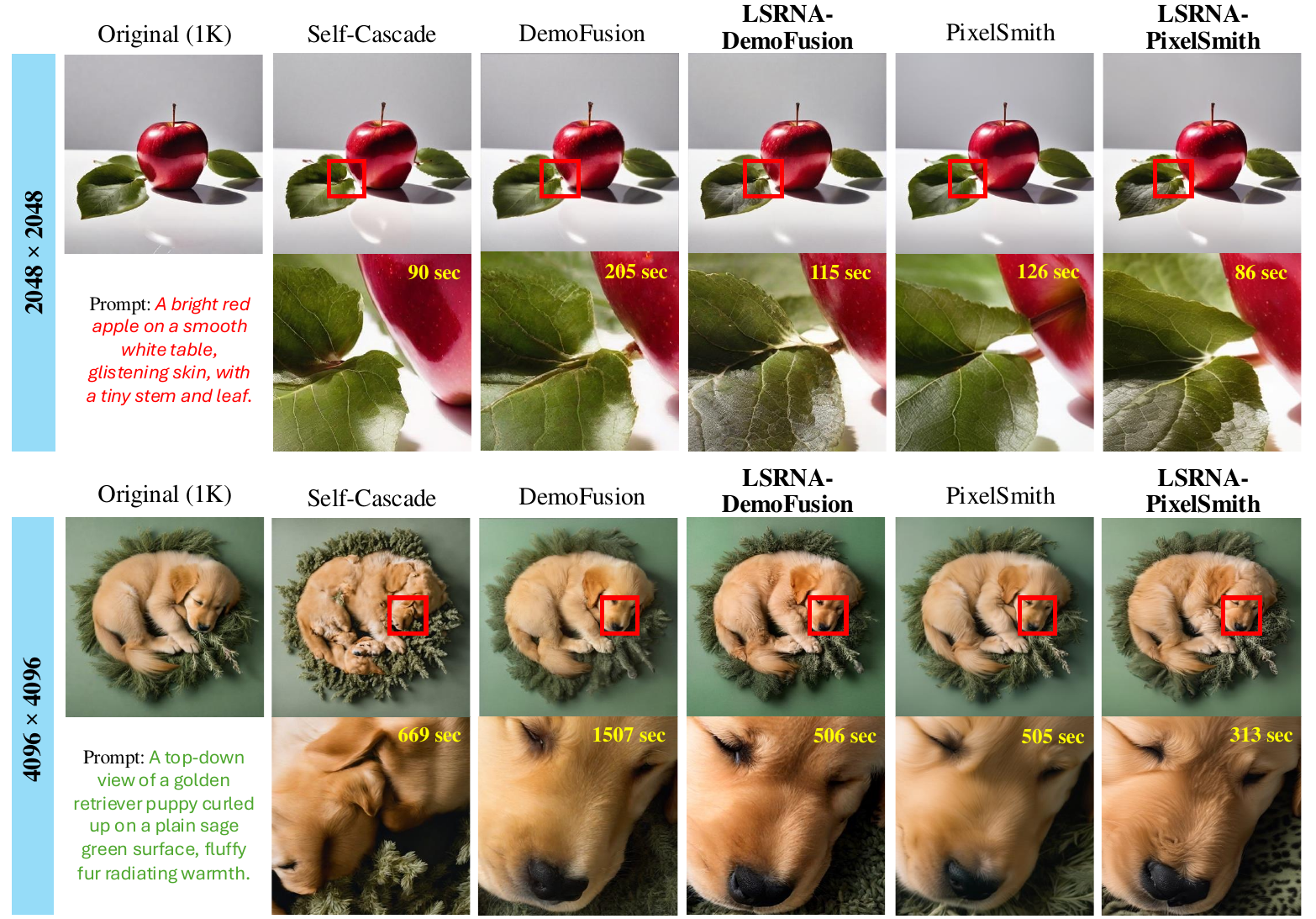}
    \caption{
    \textbf{Qualitative comparisons across reference-based methods} at 2K and 4K resolutions.
    }
    \vspace{-1em}
    \label{fig:main}
\end{figure*}

\Cref{fig:main} showcases the qualitative comparisons of three reference-based methods: DemoFusion, Pixelsmith, Self-Cascade at 2K and 4K resolutions.
DemoFusion, which utilizes latent upsampling, excels at generating fine details, as evidenced by the texture on the leaf at 2K and the intricate fur of the puppy at 4K. 
%Its latent upsampling approach enables sharper and more detailed outputs, making it highly effective for higher-resolution image generation.
In contrast, Pixelsmith, an RGB upsampling-based method, struggles to preserve detail especially at 4K resolution.
While it performs reasonably well at 2K, its reliance on RGB upsampling-based guidance results in smoother and less detailed outputs, as seen in the puppy’s fur at 4K.
Self-Cascade also performs well at 2K resolution, generating coherent and visually appealing images. 
However, at 4K resolution, it produces structurally distorted outputs despite retaining sharpness and detail. 
Self-Cascade, as a latent upsampling-based method, preserves sharpness and fine details, however, struggles to scale effectively to higher resolutions due to its inability to address manifold deviations introduced during upsampling.

Integrating DemoFusion and Pixelsmith into LSRNA framework results in noticeable improvements in detail for both methods. 
LSRNA-DemoFusion enhances DemoFusion’s already strong capability to generate fine details, producing sharper textures and more refined outputs.
Meanwhile, LSRNA-Pixelsmith shifts Pixelsmith’s guidance from RGB-based upsampling to latent-based upsampling.
This addresses  the smoothness and lack of detail inherent in RGB-based upsampling, resulting in richer and more intricate textures.
%, as seen in features like the puppy’s nose and the apple’s surface.
These results underscore the effectiveness of latent upsampling in achieving both sharper details and improved textures. 
Additional experiments in the appendix further validate the advantages of latent upsampling over RGB upsampling, highlighting LSRNA’s impact on higher-resolution image generation.

\subsection{Ablation Studies}
\begin{figure}[t]
    \centering
    \includegraphics[width=\linewidth]{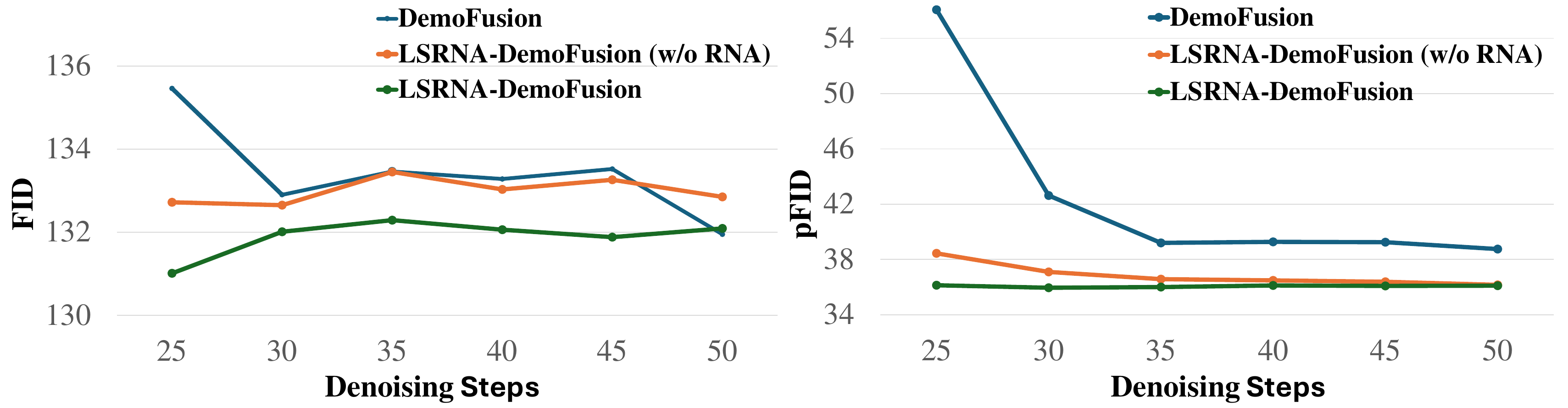}
    \vspace{-2em}
    \caption{\textbf{Ablation study of denoising steps} with DemoFusion.}
    \label{fig:denoising_quality_curve}
    \vspace{-2em}
\end{figure}

\noindent \textbf{Denoising steps and quality.}
\Cref{fig:denoising_quality_curve} plots how image quality varies with different denoising steps. 
As shown, both FID and pFID improve in the order DemoFusion → LSRNA-DemoFusion (w/o RNA) → LSRNA-DemoFusion, with our framework maintaining stable trends across denoising steps.
A key reason for this stability is that our LSR module places the upsampled latent closer to the HR manifold, allowing fewer denoising steps to refine the image with minimal performance loss. 
In contrast, DemoFusion’s less aligned upsampled latent requires full noise injection (\ie, 50 steps) to align with the manifold.
Although FID and pFID are similar across denoising steps for LSRNA, our inspection shows that steps below 30 produce unstable quality, leading us to adopt 30 steps.

\begin{figure*}[t]
    \centering
    \includegraphics[width=\linewidth]{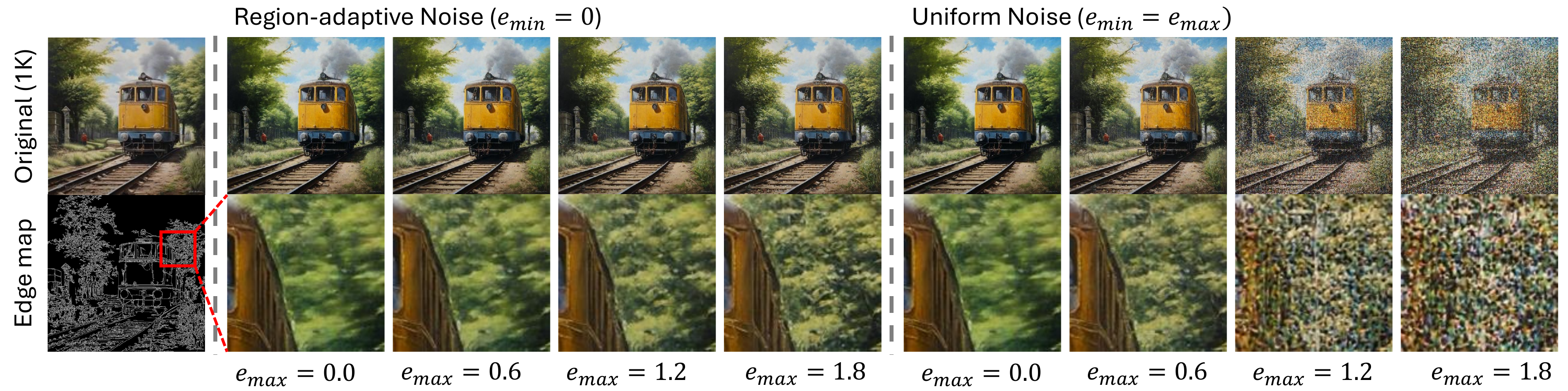}
    \vspace{-1.5em}
    \caption{
        \textbf{Qualitative comparison of Region-wise Noise Addition (RNA) and Uniform Noise Addition (UNA).} 
        RNA (left) effectively preserves low-frequency areas while enhancing high-frequency detail and texture.
        In contrast, UNA (right) introduce artifacts across various regions at higher noise levels.
    }
    \vspace{-1em}
    \label{fig:rna}
\end{figure*}

\noindent\textbf{Effectiveness of LSR \& RNA.}
\Cref{tab:lsrna} compares DemoFusion and its ablations with LSR and RNA. 
Integrating LSR reduces inference time with minimal performance degradation, demonstrating that it effectively provides high-quality latent guidance by aligning low-resolution representations with the higher-resolution manifold. 
Adding RNA further refines the output by enhancing finer details in high-frequency regions, leading to improved performance especially on the patch-based metrics.

\begin{table}[]
\caption{
\textbf{LSR \& RNA ablation} on OpenImages-Valid ($\times$9) with DemoFusion. The best results marked in \textbf{bold}.
}
\vspace{-2mm}
% \resizebox{\linewidth}{!}{%%
% \renewcommand{\arraystretch}{1.25}
% \begin{tabular}{c|cccccc}
% \toprule
%             & FID (↓)         & KID (↓)         & pFID (↓)       & pKID (↓)        & IS (↑)         & CLIP (↑)       \\ \hline
% DemoFusion   & 132.90          & 0.0069          & 42.63          & 0.0103          & 27.43          & 0.305          \\
% w/ LSR      & 132.65          & 0.0065          & 37.10          & \textbf{0.0057} & 27.84          & \textbf{0.310} \\
% w/ LSR, RNA & \textbf{132.01} & \textbf{0.0053} & \textbf{35.95} & \textbf{0.0057} & \textbf{29.87} & \textbf{0.310}
%           \\ \bottomrule
% \end{tabular}
% }%%
\resizebox{\linewidth}{!}{%%
\renewcommand{\arraystretch}{1.5}
\begin{tabular}{l|cccc|c}
\toprule
                           & FID (↓)         & KID (↓)         & pFID (↓)       & pKID (↓)        & Time (sec) \\ \hline
DemoFusion                 & \textbf{131.95} & 0.0064          & 38.75          & 0.0075          & 660        \\ \cdashline{1-6}
LSRNA-DemoFusion (w/o RNA) & 132.65          & 0.0065          & 37.10          & \textbf{0.0057} & 272        \\
LSRNA-DemoFusion           & 132.01          & \textbf{0.0053} & \textbf{35.95} & \textbf{0.0057} & 272        \\ \bottomrule
\end{tabular}
}%%
\vspace{-4mm}
\label{tab:lsrna}
\end{table}

\noindent\textbf{Importance of Region-Adaptiveness in RNA.}
As shown in \Cref{fig:rna}, Uniform Noise Addition (UNA) introduce artifacts across various regions, particularly at higher noise levels.
By employing a Region-wise Noise Addition (RNA), we can effectively preserve low-frequency areas while facilitating high-frequency detail and texture.
%Quantitative experiments on RNA and UNA are presented in the appendix.

\noindent\textbf{Impact of RNA strength.}
%\Cref{tab:rna} demonstrates the effect of \( e_{\text{max}} \) (when \( e_{\text{min}}=0 \)), which controls the strength of RNA across various resolutions.
\Cref{tab:rna} demonstrates the effect of the strength of RNA (\ie, \( e_{\text{max}} \)) across various resolutions.
The results indicate that \( e_{\text{max}} = 1.2 \) yields the lowest pFID across all scales, suggesting that it effectively balances detail enhancement and content over-generation. 
Therefore, we select \([e_{\text{min}}, e_{\text{max}}] = [0.0, 1.2]\) for DemoFusion. 
A similar ablation study for Pixelsmith is presented in the appendix, where we adopt \([e_{\text{min}}, e_{\text{max}}] = [0.4, 0.8]\).

\vspace{-1mm}
\section{Limitation}
While RNA demonstrates robustness across different denoising steps (as seen in \Cref{fig:denoising_quality_curve}), its optimal strength varies depending on the integrated method (\eg, DemoFusion or Pixelsmith) and the predefined noise schedule. 
This is because the guidance latent interacts with each method’s inherent denoising process during the second diffusion stage.
Moreover, although our approach learns to map onto the high-resolution manifold, the generation ability of LSRNA remains inherently limited by the capacity of the pretrained LDM.
\begin{table}[]
\caption{
\textbf{Ablation study on RNA strength} using OpenImages-Valid with DemoFusion.
The best results are marked in \textbf{bold}.
$e_{min}$ is set to 0.
}
\vspace{-1mm}
\resizebox{\linewidth}{!}{%%
\renewcommand{\arraystretch}{1.50}
\begin{tabular}{ccc|ccc|ccc}
\toprule
\multicolumn{3}{c|}{\textbf{$\times$4} (2048$\times$2048)}                                              & \multicolumn{3}{c|}{\textbf{$\times$9} (3072$\times$3072)}                                              & \multicolumn{3}{c}{\textbf{$\times$16} (4096$\times$4096)}                                               \\ \hline
\multicolumn{1}{c|}{$e_{max}$}         & FID (↓)             & pFID (↓)          & \multicolumn{1}{c|}{$e_{max}$}         & FID (↓)           & pFID (↓)          & \multicolumn{1}{c|}{$e_{max}$}         & FID (↓)            & pFID (↓)          \\ \hline
\multicolumn{1}{c|}{0.0}          & 132.27          & 53.40          & \multicolumn{1}{c|}{0.0}          & 132.65          & 37.10          & \multicolumn{1}{c|}{0.0}          & 135.00          & 34.00          \\ \cdashline{1-9}
\multicolumn{1}{c|}{0.8}          & 131.87          & 53.03          & \multicolumn{1}{c|}{0.8}          & 132.68          & 36.40          & \multicolumn{1}{c|}{0.8}          & 135.29          & 33.62          \\
\multicolumn{1}{c|}{1.0}          & 132.13          & 53.28          & \multicolumn{1}{c|}{1.0}          & 132.43          & 36.12          & \multicolumn{1}{c|}{1.0}          & 135.09          & 33.64          \\
\multicolumn{1}{c|}{\textbf{1.2}} & 131.47          & \textbf{52.96} & \multicolumn{1}{c|}{\textbf{1.2}} & 132.01          & \textbf{35.95} & \multicolumn{1}{c|}{\textbf{1.2}} & 134.90          & \textbf{33.35} \\
\multicolumn{1}{c|}{1.4}          & \textbf{131.44} & 53.12          & \multicolumn{1}{c|}{1.4}          & \textbf{131.73} & 36.04          & \multicolumn{1}{c|}{1.4}          & \textbf{134.71} & 33.77          \\ \bottomrule
\end{tabular}
}%%
\vspace{-4.5mm}
\label{tab:rna}
\end{table}
\vspace{-2mm}
\section{Conclusion}
In this paper, we address the challenges in higher-resolution image generation with diffusion models by identifying manifold deviation in latent upsampling and insufficient texture detail in RGB upsampling.
To tackle these issues, we introduce LSR, which aligns low-resolution latent representations with the higher-resolution manifold, and RNA, which adaptively enhances finer details in high-frequency regions.
We then propose LSRNA, a novel framework that combines LSR and RNA.
By incorporating LSRNA into both latent and RGB upsampling-based methods, we demonstrate its ability to outperform state-of-the-art methods across various resolutions and metrics. 
Our experimental results highlight the ability of LSRNA, pushing the boundaries of higher-resolution image generation.
\vspace{-2mm}
\section*{Acknowledgement}
This research was supported and funded by the Artificial Intelligence Graduate School Program under Grant 2020-0-01361 and by the Institute of Information \& Communications Technology Planning \& Evaluation (IITP) grant funded by the Korea government (MSIT) under Grant 2022-0-00124.

%------------------------------------------------------
\clearpage
\renewcommand{\thesection}{\Alph{section}}
\renewcommand{\thefigure}{\Alph{figure}}
\renewcommand{\thetable}{\Alph{table}}
\setcounter{section}{0}
\setcounter{table}{0}
\setcounter{figure}{0}

\setcounter{page}{1}
\maketitlesupplementary

\section{Importance of Latent Space Upsampling}
One of our key findings is that for reference-based higher-resolution image generation methods~\cite{du2024demofusion,tragakis2024one,guo2024make}, quality of output images differs significantly depending on whether the reference is upsampled in RGB space or latent space. 
We hypothesize that upsampling within the latent space plays a crucial role in preserving the sharpness and detail essential for higher-resolution image generation.
In this section, we provide additional qualitative and quantitative experimental results to support our hypothesis.

\subsection{Setting}
We define several RGB upsampling variants of the existing models DemoFusion~\cite{du2024demofusion} and Pixelsmith~\cite{tragakis2024one} by modifying their reference upsampling strategies. 
First, we introduce DemoFusion-rgbBic for the DemoFusion model, where the reference is upsampled in RGB space using bicubic interpolation. 
For the Pixelsmith model, we define Pixelsmith-rgbLanc, which employs Lanczos interpolation~\cite{lanczos1964evaluation} in RGB space for reference upsampling. 
Pixelsmith-rgbLanc corresponds to the original Pixelsmith model.

Building on these, we further define variants that perform super-resolution (SR) in RGB space using a separate SR network, namely DemoFusion-rgbSR and Pixelsmith-rgbSR. 
The SR network shares the same architecture and training settings as our LSR module (detailed in \Cref{sec:LSR_training}), with the input and output channels is set to 3 to process RGB images.

In contrast to these variants, LSRNA-DemoFusion and LSRNA-Pixelsmith utilize latent space upsampling through our proposed LSRNA framework. 
While the original DemoFusion also performs bicubic upsampling in latent space and demonstrates strengths in preserving detail, we instead demonstrate the effectiveness of latent upsampling within the LSRNA framework.

\subsection{Analysis}
Our qualitative results, presented in Figures~\ref{fig:appendix_latent_importance1} and \ref{fig:appendix_latent_importance2} for 16$\times$ resolution and Figures~\ref{fig:appendix_latent_importance3} and \ref{fig:appendix_latent_importance4} for 64$\times$ resolution, demonstrate consistent trends when comparing RGB upsampling variants to latent upsampling using our LSRNA framework. 
Specifically, RGB upsampling methods, whether based on interpolation or super-resolution, produce smoother images that lack fine details. In contrast, latent upsampling yields sharper and more detailed results.

Our quantitative results in Table~\ref{tab:appendix_latent_importance} further confirm these observations that latent upsampling approaches (\ie, LSRNA-DemoFusion and LSRNA-Pixelsmith) consistently outperform their RGB upsampling counterparts.
The improvements are particularly evident in patch-based metrics like pFID and pKID, which focus on capturing finer details.
These results underscore the critical role of latent space upsampling in enhancing local detail fidelity and textures.

We attribute these findings to the representational characteristics of the latent space. 
Unlike RGB space, the latent space encodes image features in a compressed form, capturing high-level information. 
Upsampling within this domain likely leverages these representations to better preserve fine details and sharpness. 
Conversely, RGB space upsampling is constrained by the raw pixel-level representation, which acts as a bottleneck and hinders the preservation of details and textures. 
Further exploration is needed to fully understand the underlying reasons behind these differences.
\begin{table}[]
\caption{
\textbf{RGB vs. Latent Space Upsampling} on OpenImages-Valid ($\times$9).
The best results marked in \textbf{bold}.
}
\resizebox{\linewidth}{!}{%%
\renewcommand{\arraystretch}{1.5}
\begin{tabular}{l|cccccc}
\toprule
                   & FID (↓)         & KID (↓)         & pFID (↓)       & pKID (↓)               \\ \bottomrule \toprule
DemoFusion-rgbBic  & 134.56          & 0.0084          & 37.44          & 0.0062                    \\
DemoFusion-rgbSR   & 134.55          & 0.0093          & 37.35          & 0.0061                    \\ \cdashline{1-5}
LSRNA-DemoFusion   & \textbf{132.01} & \textbf{0.0053} & \textbf{35.95} & \textbf{0.0057}  \\ \midrule
Pixelsmith-rgbLanc & 134.31          & 0.0095          & 40.64          & 0.0084           \\
Pixelsmith-rgbSR   & 134.34          & 0.0102          & 44.41          & 0.0110           \\ \cdashline{1-5}
LSRNA-Pixelsmith   & \textbf{132.17} & \textbf{0.0077} & \textbf{36.71} & \textbf{0.0057}           \\ \bottomrule
\end{tabular}
}%%
\vspace{-1.5em}
\label{tab:appendix_latent_importance}
\end{table}
\section{Experimental Details}
\subsection{Comparison}
To ensure a fair comparison, all experiments across the main text and appendix are conducted with a consistent setup unless otherwise specified.
This includes the use of a fixed random seed, tiled decoding, negative prompts derived from DemoFusion, a guidance scale of 7.5, xFormers~\cite{lefaudeux2022xformers} enabled with float16 precision, and FreeU~\cite{si2024freeu} disabled. 
In addition, unnecessary visualization code like intermediate image reconstruction is disabled for runtime measurement.

We employ a DDIM~\cite{song2020denoising} scheduler, with the $\eta$ parameter set to 0 for reference-based methods, and $\eta=1$ for non-reference-based methods, as $\eta=0$ leads to noticeable degradation in quantitative performance for the latter.
Existing methods use 50 DDIM steps for higher-resolution generation process, while LSRNA uses 30 steps.

\subsection{Patch-Based Metrics}
Conventional metrics like FID~\cite{heusel2017gans} and KID~\cite{binkowski2018demystifying} involve resizing images to $299^2$, which can lead to a loss of high-resolution details. 
To address this issue, inspired by Anyres-GAN~\cite{chai2022anyresolution}, we adopt patch-based metrics (pFID and pKID) that focus on local details and textures, which are critical for evaluating high-resolution image generation.

The patch-based metrics are computed by first cropping the original ground truth images to match the aspect ratio of the generated images, followed by resizing them to the same resolution using Lanczos interpolation.
Next, 1K-sized patches are cropped from both the generated and ground truth images at 50,000 randomly selected locations. 
For a fair comparison, a fixed random seed is used to maintain consistency in crop locations both between generated and ground truth images and across different generation methods.
These extracted patches are then used to calculate the FID and KID metrics, referred to as pFID and pKID.
\vspace{-1mm}
\section{LSR Training Details}
\subsection{Data Preparation}
To prepare LR-HR latent pairs for training the LSR module, we leverage the real-world dataset~\cite{kuznetsova2020open} to obtain ground-truth HR RGB images. 
We construct training pairs in two steps: (i) downsampling the HR RGB images to generate LR RGB images, and (ii) encoding the HR and LR RGB images independently using a pretrained encoder $\mathcal{E}$.

\vspace{1mm}
\noindent \textbf{Bridging the Domain Gap.}
To address the domain gap between training and inference, we simply adopt bicubic degradation over complex real-world degradations~\cite{zhang2021designing, wang2021real} in step (i).
This choice aligns with the inference scenario, where LR images (decoded from LR latents) typically exhibit minimal noise or artifacts. 
Bicubic degradation avoids the noise or artifacts introduced by real-world degradations while significantly reducing preprocessing time.

In step (ii), directly downsampling HR latents to create LR latents is avoided, as it can cause inconsistencies within the latent manifold. 
Instead, our approach ensures that LR and HR latents are encoded separately, preserving consistency within their respective manifolds.

\vspace{1mm}
\noindent \textbf{Multi-Scale Preparation.}
To enable multi-scale training for the LSR module, we filter ground-truth HR RGB images with a minimum resolution of 1440 pixels in both height and width. 
For each HR image, a crop size is randomly selected between $[1056, 1440]$ in multiples of 96 (chosen to align with the downscaling factors and the encoder's compression ratio of 8). 
The HR image is then divided into non-overlapping patches of the selected crop size, forming HR RGB patches. 
Each HR patch is subsequently encoded into the latent space.

To create LR counterparts, each HR RGB patch is downscaled by factors of $\times$2, $\times$3, and $\times$4. 
The resulting LR RGB patches are then encoded using the same encoder to generate LR latent representations. 
Our data preparation process results in a training dataset comprising a total of 4.7M LR-HR latent pairs with diverse scale.

\subsection{Batch Construction}
Each LR-HR latent pair has varying sizes, necessitating alignment of the spatial dimensions between the input LR latent and target HR latent for batching.
During data-loading, we further randomly crop the LR latents to a fixed size of 32×32 pixels.
From the corresponding HR latents, 4096 pixels within the cropped region are randomly sampled to serve as the ground truth.
Additionally, we avoid data augmentation techniques such as horizontal and vertical flips, as they can lead to deviations in the latent space manifold.
Our batching strategy not only ensures efficient training but also enables the LSR to learn mappings from LR to HR latent representations across multiple scales.

\subsection{Training}
\label{sec:LSR_training}
\begin{table}[]
\caption{
\textbf{Quantitative comparison of image generation results by LSR training variants} on OpenImages-Valid ($\times$9).
All training is conducted on a single NVIDIA Tesla V100-SXM2 GPU, using SwinIR~\cite{liang2021swinir} and RCAN~\cite{zhang2018image} as backbones with LIIF~\cite{chen2021learning} as the upsampler. 
Based on a balance of training efficiency and performance, we adopt the v1 configuration.
}
\vspace{-2mm}
\resizebox{\linewidth}{!}{%%
\renewcommand{\arraystretch}{1.5}
\begin{tabular}{c|c|c|c|c}
\toprule
LSRNA-DemoFusion                      & \textbf{v1 (adpoted)}     & v2                       & v3                       & v4                       \\ \bottomrule \toprule
Params                & 1.29M                    & 1.29M                    & 1.29M                    & 15.64M                   \\
Backbone              & SwinIR (Light)           & SwinIR (Light)           & SwinIR (Light)           & RCAN                     \\ \cdashline{1-5}
Initial learning rate & $2 \times 10^{-4}$ & $2 \times 10^{-4}$ & $1 \times 10^{-4}$ & $2 \times 10^{-4}$ \\
Batch size            & 32                       & 32                       & 16                       & 32                       \\
Training iteration    & 200K                     & 1000K                    & 200K                     & 200K                     \\
Training time         & 26h                      & 129h                     & 15h                      & 26h                      \\ \cdashline{1-5}
FID (↓)                  & 134.84                   & 134.90                    & 134.28                   & 134.25                   \\
KID (↓)                  & 0.0077                   & 0.0077                   & 0.0077                   & 0.0074                   \\
pFID (↓)                 & 33.47                    & 33.35                    & 33.75                    & 34.34                    \\
pKID (↓)                 & 0.0073                   & 0.0072                   & 0.0074                   & 0.0076                   \\ \bottomrule
\end{tabular}
}%%
\vspace{-4mm}
\label{tab:appendix_lsr_setting}
\end{table}
We adopt SwinIR~\cite{liang2021swinir} as the backbone for the LSR and LIIF~\cite{chen2021learning} as the upsampler, modifying both input and output channel dimensions to match the latent space dimension of 4.
The optimizer used is Adam~\cite{kingma2014adam} with an initial learning rate of $2 \times 10^{-4}$, scheduled with cosine annealing.
Training is performed over 200K iterations with a batch size of 32. The loss function is defined by an $L_1$ loss in the latent space. 
We leave the exploration of other loss functions (\eg, perceptual loss~\cite{zhang2018unreasonable}) for future work.

Quantitative results of image generation by various LSR training settings are presented in \Cref{tab:appendix_lsr_setting}, while also demonstrating the efficiency of the LSR training.
Although the LSR is trained on the paired dataset constructed with relatively sparse downscaling factors, it can generalize to arbitrary scaling factors during inference, enabled by our multi-scale training scheme and the generalization capability of LIIF.
Our motivation for using LIIF upsampler instead of traditional fixed-scale upsampler~\cite{shi2016real} lies in its ability to handle arbitrary resolutions with a single LSR module trained once.
\begin{figure*}[t]
    \centering
    \includegraphics[width=\linewidth]{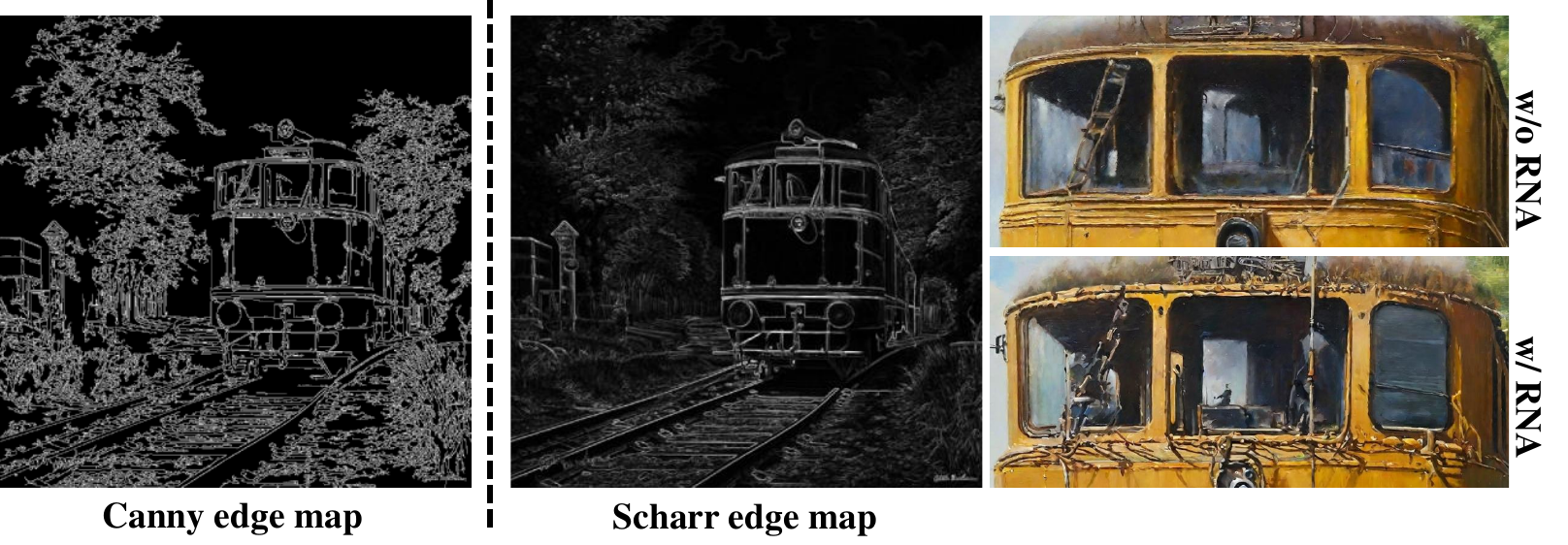}
    %\vspace{-6mm}
    \caption{\textbf{Qualitative results of RNA using Scharr edge map.}}
    %\vspace{-3mm}
    \label{fig:scharr}
\end{figure*}
\section{Edge Detection for RNA}
RNA is designed to adaptively add Gaussian noise to specific areas of the upsampled reference latent, focusing on detail-critical regions (\ie, high-frequency regions). Our intuition behind RNA is that introducing irregularities in regions that would otherwise remain flat prompts the diffusion model to synthesize new details in those regions.

To identify these areas, we consider using edge detection algorithms. 
However, we find that common edge detectors such as Scharr~\cite{scharr2000optimal}, LoG~\cite{marr1980theory}, and Gabor~\cite{gabor1946theory}, which primarily focus on precise object boundaries, tend to produce artifacts such as overgeneration around contours or jagged contours when used as the basis for RNA. 
We present qualitative results in \Cref{fig:scharr} using the Scharr edge map, which is known for effectively capturing weak edges.
As shown, while strong edges (\eg, on the train’s window) are sparsely detected, weak edges (\eg, on the tree) appear with excessively low intensity. This results in artifacts or over-enhanced details around strong edges when RNA is applied.

To address this, we adopt Canny edge detection~\cite{canny1986computational}, which allows us to prioritize weak edges by adjusting the lower and upper thresholds.
By doing so, we can detect detailed regions rather than strictly connected edge lines, as demonstrated in \Cref{fig:scharr}.
This region-based detection allows RNA to enhance local details effectively without introducing edge-based artifacts.
\begin{table}[]
%\vspace{-1em}
\centering
% \scriptsize
% \setlength{\tabcolsep}{10pt}
\caption{
\textbf{Quantitative comparison of image generation results by Canny edge detection thresholds} on OpenImages-Valid ($\times9$) with DemoFusion. $e_{max}$ is set to 1.2 (default).
}
\resizebox{\linewidth}{!}{%%
\renewcommand{\arraystretch}{1.0}
\begin{tabular}{cc|cccc}
\hline
lower & upper & FID (↓)   & KID (↓)   & pFID (↓) & pKID (↓)  \\ \bottomrule \toprule
0         & 255       & 132.01 & 0.0053 & 35.95 & 0.0057 \\
30        & 180       & 132.18 & 0.0055 & 36.01 & 0.0057 \\
50        & 200       & 132.54 & 0.0055 & 36.09 & 0.0057 \\
60        & 150       & 132.19 & 0.0055 & 36.12 & 0.0057 \\ \hline
\end{tabular}
}%%
\label{tab:rebuttal_canny_threshold}
\end{table}
\begin{figure*}[t]
    \centering
    \includegraphics[width=\linewidth]{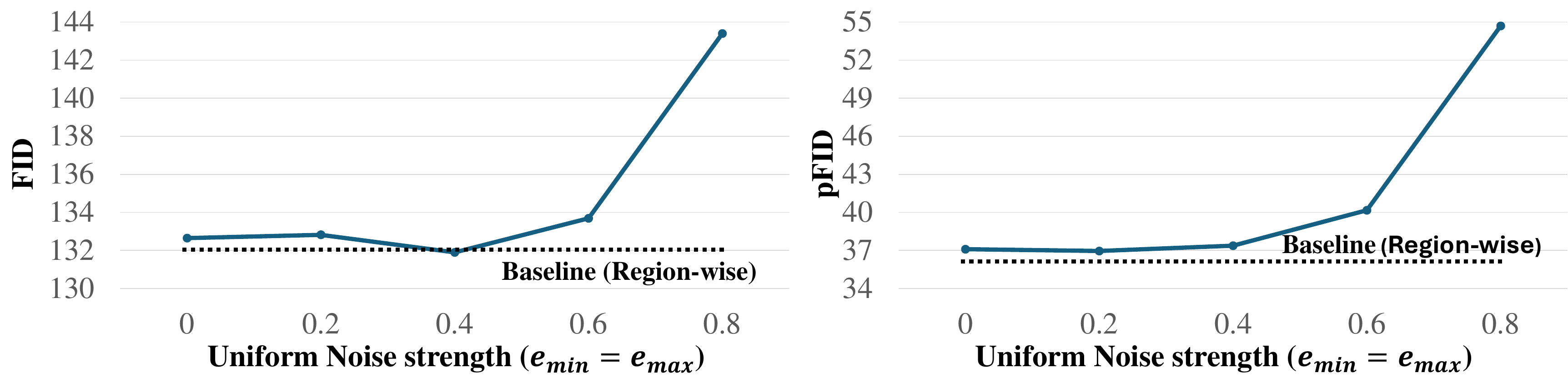}
    \caption{\textbf{Ablation study of UNA (Uniform Noise Addition) strength} on OpenImages-Valid ($\times9$) with DemoFusion.
    Dotted line shows our default RNA setting ($e_{min}=0$ and $e_{max}=1.2$).}
    \label{fig:rebuttal_uniform_rna}
\end{figure*}
\begin{table}[]
% \scriptsize
\caption{
\textbf{Pixel-wise difference based on RNA strength.} Differences are computed after applying RNA compared to no RNA. Histogram matching is applied before computing differences. The scene from the main Figure 7 is used.
}
\resizebox{\linewidth}{!}{%%
\renewcommand{\arraystretch}{1.2}
\begin{tabular}{cc|ccc|cc|ccc}
\hline
$e_{min}$ & $e_{max}$ & Non-edge & Edge  & Gap           & $e_{min}$ & $e_{max}$ & Non-edge & Edge  & \multicolumn{1}{l}{Gap} \\ \bottomrule \toprule
0      & 0.6    & 1.36     & 3.44  & \textbf{2.09} & 0.6    & 0.6    & 6.36     & 9.61  & 3.25                    \\
0      & 1.2    & 2.65     & 6.57  & \textbf{3.92} & 1.2    & 1.2    & 25.95    & 29.06 & 3.11                    \\
0      & 1.8    & 4.01     & 10.27 & \textbf{6.26} & 1.8    & 1.8    & 39.12    & 39.81 & 0.7                     \\ \hline
\end{tabular}
}%%
\label{tab:rebuttal_pixel_difference}
\end{table}

\section{Robustness of RNA}
We demonstrate in \Cref{tab:rebuttal_canny_threshold} that generation performance is robust to variations in the Canny's thresholds, as long as they are set to prioritize weak edges.
In our implementation, we use a lower threshold of 0 and an upper threshold of 255. 
We further evaluate the robustness of RNA through additional experiments: \Cref{fig:rebuttal_uniform_rna} indicates that RNA is quantitatively better than UNA (Uniform Noise Addition), while \Cref{tab:rebuttal_pixel_difference} shows that RNA indeed enhances details where the edge map is activated.
\section{Additional Ablation Studies}
\begin{table}[]
\caption{
\textbf{LSR \& RNA ablation} on OpenImages-Valid ($\times$9).
The best results marked in \textbf{bold}.
}
\resizebox{\linewidth}{!}{%%
\renewcommand{\arraystretch}{1.6}
\begin{tabular}{l|cccc|c}
\toprule
                           & FID (↓)         & KID (↓)         & pFID (↓)       & pKID (↓)        & Time (sec) \\ \bottomrule \toprule
DemoFusion                 & \textbf{131.95} & 0.0064          & 38.75          & 0.0075          & 660        \\ \cdashline{1-6}
LSRNA-DemoFusion (w/o RNA) & 132.65          & 0.0065          & 37.10          & \textbf{0.0057} & 272        \\
LSRNA-DemoFusion           & 132.01          & \textbf{0.0053} & \textbf{35.95} & \textbf{0.0057} & 272        \\ \midrule
Pixelsmith                 & 134.31          & 0.0095          & 40.64          & 0.0084          & 289        \\ \cdashline{1-6}
Pixelsmith-latentBic     & 142.23          & 0.0155          & 67.91          & 0.0275          & 291        \\
LSRNA-Pixelsmith (w/o RNA) & 137.71          & 0.0116          & 46.45          & 0.0112          & 181        \\
LSRNA-Pixelsmith           & \textbf{132.17} & \textbf{0.0077} & \textbf{36.71} & \textbf{0.0057} & 182        \\ \bottomrule
\end{tabular}
}%%
\label{tab:appendix_lsrna}
\end{table}
\subsection{Effectiveness of LSR \& RNA}
We provide additional quantitative results to assess the impact of the LSR and RNA modules on both the DemoFusion and Pixelsmith models. 
For the original Pixelsmith, which performs upsampling in the RGB space unlike DemoFusion, we introduce an additional variant called \textit{Pixelsmith-latentBic}. 
This variant replaces the original RGB space upsampling with bicubic interpolation in the latent space.

The results are summarized in \Cref{tab:appendix_lsrna}. 
For DemoFusion, incorporating the LSR module enhances performance by providing high-quality latent guidance, improving image generation quality even with fewer denoising steps and without progressive upscaling. 
The addition of the RNA module further boosts performance by enriching finer details and textures in the generated images. 

In case of Pixelsmith, replacing the original RGB upsampling with latent upsampling (\ie, Pixelsmith-latentBic) leads to significant performance degradation.
However, applying the LSR module to perform super-resolution in the latent space leads to a noticeable improvement in performance. 
The RNA module further improves the results and ultimately surpasses the performance of the original model, demonstrating the adaptability and effectiveness of our LSR and RNA modules.

\subsection{Impact of RNA Strength}
\begin{table}[]
\caption{
\textbf{Ablation Study on RNA strength with Pixelsmith} on OpenImages-Valid ($\times$9).
The best results marked in \textbf{bold}.
}
\resizebox{\linewidth}{!}{%%
\renewcommand{\arraystretch}{1.2}
\begin{tabular}{cccc|cccc}
\toprule
$e_{min}$ & $e_{max}$ & FID (↓)   & pFID (↓) & $e_{min}$ & $e_{max}$ & FID (↓)            & pFID (↓)          \\ \bottomrule \toprule
0.0       & 0.0       & 137.71 & 46.45 & 0.2       & 1.2       & 133.67          & 37.76          \\
0.0       & 1.2       & 135.15 & 40.45 & 0.2       & 1.4       & 133.72          & 38.51          \\
0.0       & 1.4       & 134.32 & 38.75 & 0.4       & 0.6       & 132.18          & 37.3           \\
0.0       & 1.6       & 134.34 & 38.74 & 0.4       & 0.8       & \textbf{132.17} & \textbf{36.71} \\
0.2       & 1.0       & 133.86 & 38.51 & 0.4       & 1.0       & 132.29          & 36.95          \\ \bottomrule
\end{tabular}
}%%
\label{tab:appendix_rna}
\end{table}

Building on the RNA strength tuning results for LSRNA-DemoFusion presented in the main text, we further evaluate the impact of RNA strength on LSRNA-Pixelsmith, as shown in \Cref{tab:appendix_rna}.
While LSRNA-DemoFusion achieves optimal performance with $e_{min} = 0$ (and $e_{max} = 1.2$), LSRNA-Pixelsmith performs best with $e_{min} = 0.4$ and $e_{max} = 0.8$. 
This difference in optimal RNA strength arises from the distinct roles played by the reference latent in the high-resolution generation process of each model. 
LSRNA-Pixelsmith likely requires a higher $e_{min}$ to ensure effective noise injection into the reference latent.
The RNA strength determined from this validation process is consistently applied across all other experiments including those in the main text.
\begin{figure*}[t]
    \centering
    \includegraphics[width=0.87\linewidth]{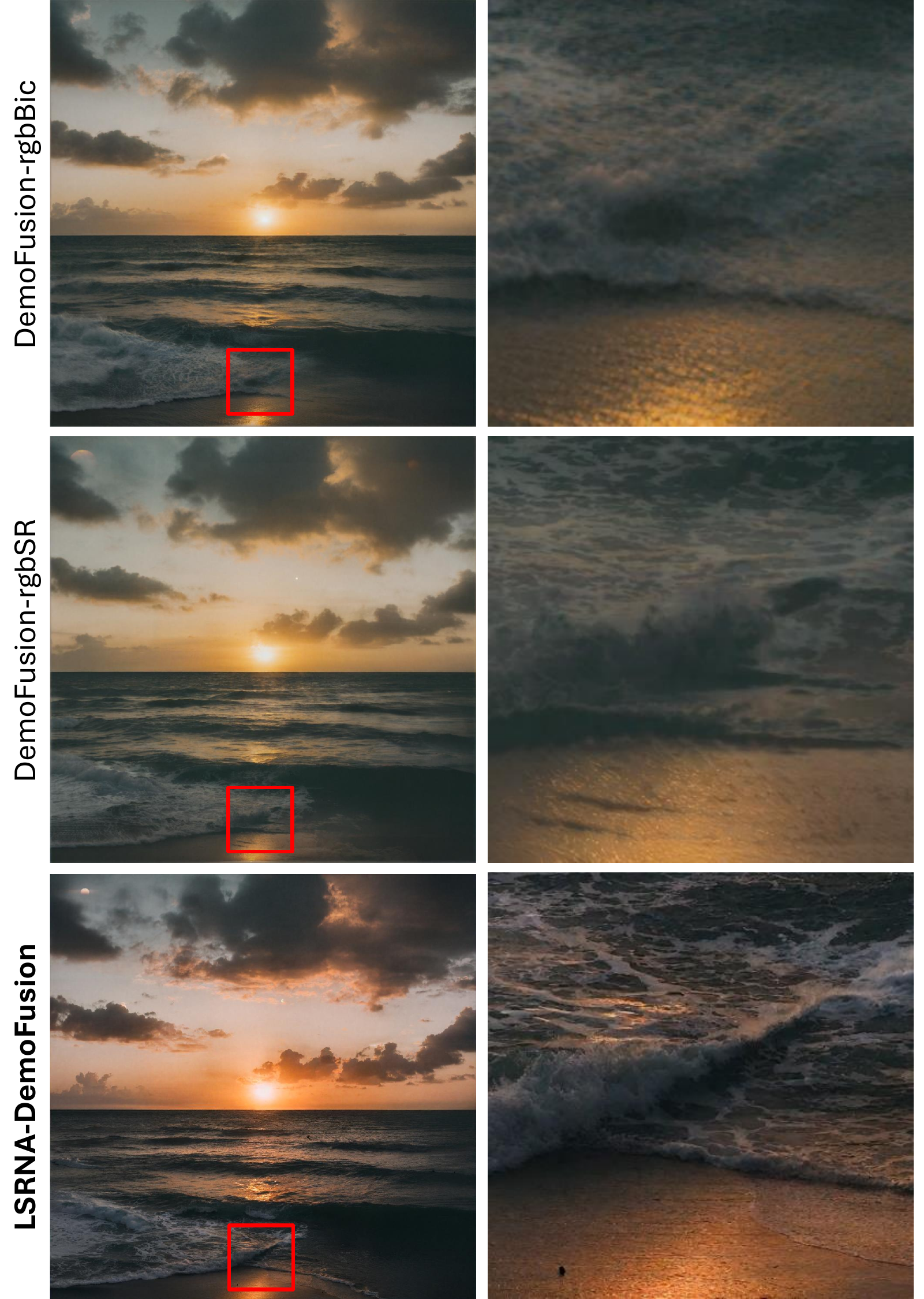}
    % \vspace{-2em}
    \caption{
    \textbf{RGB vs. Latent Space Upsampling for DemoFusion on 16$\times$.}
    Prompt used is "the sun is setting over the ocean on a cloudy day".
    Best viewed \textbf{ZOOMED-IN}.
    }
    \label{fig:appendix_latent_importance1}
\end{figure*}
\begin{figure*}[t]
    \centering
    \includegraphics[width=0.87\linewidth]{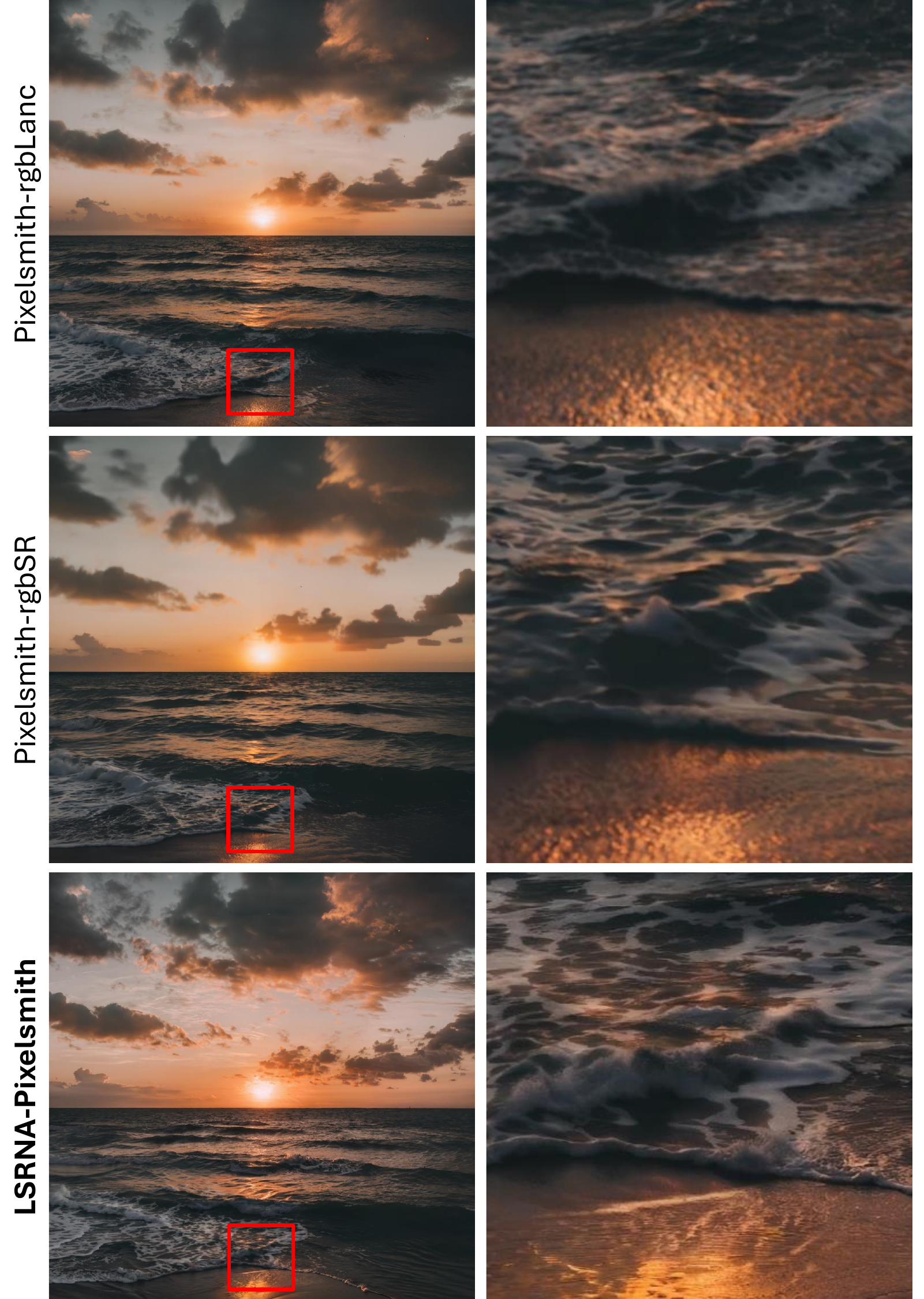}
    % \vspace{-2em}
    \caption{
    \textbf{RGB vs. Latent Space Upsampling for Pixelsmith on 16$\times$.}
    Prompt used is "the sun is setting over the ocean on a cloudy day".
    Best viewed \textbf{ZOOMED-IN}.
    }
    \label{fig:appendix_latent_importance2}
\end{figure*}
\begin{figure*}[t]
    \centering
    \includegraphics[width=0.87\linewidth]{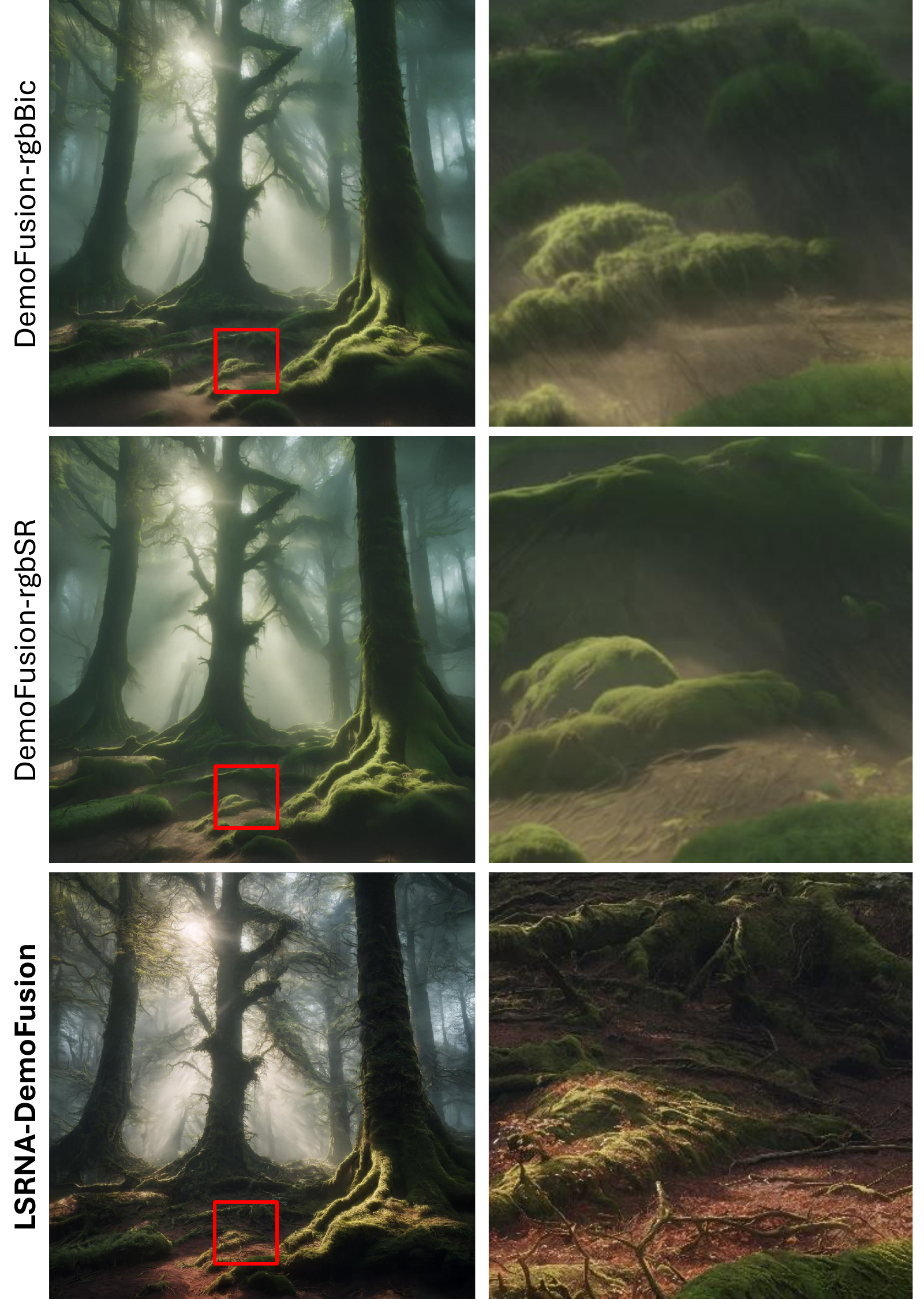}
    % \vspace{-2em}
    \caption{
    \textbf{RGB vs. Latent Space Upsampling for DemoFusion on 64$\times$.}
    Prompt used is "A mysterious forest with tall, ancient trees and beams of sunlight filtering through the mist, detailed moss-covered roots, 8k".
    Best viewed \textbf{ZOOMED-IN}.
    }
    \label{fig:appendix_latent_importance3}
\end{figure*}
\begin{figure*}[t]
    \centering
    \includegraphics[width=0.87\linewidth]{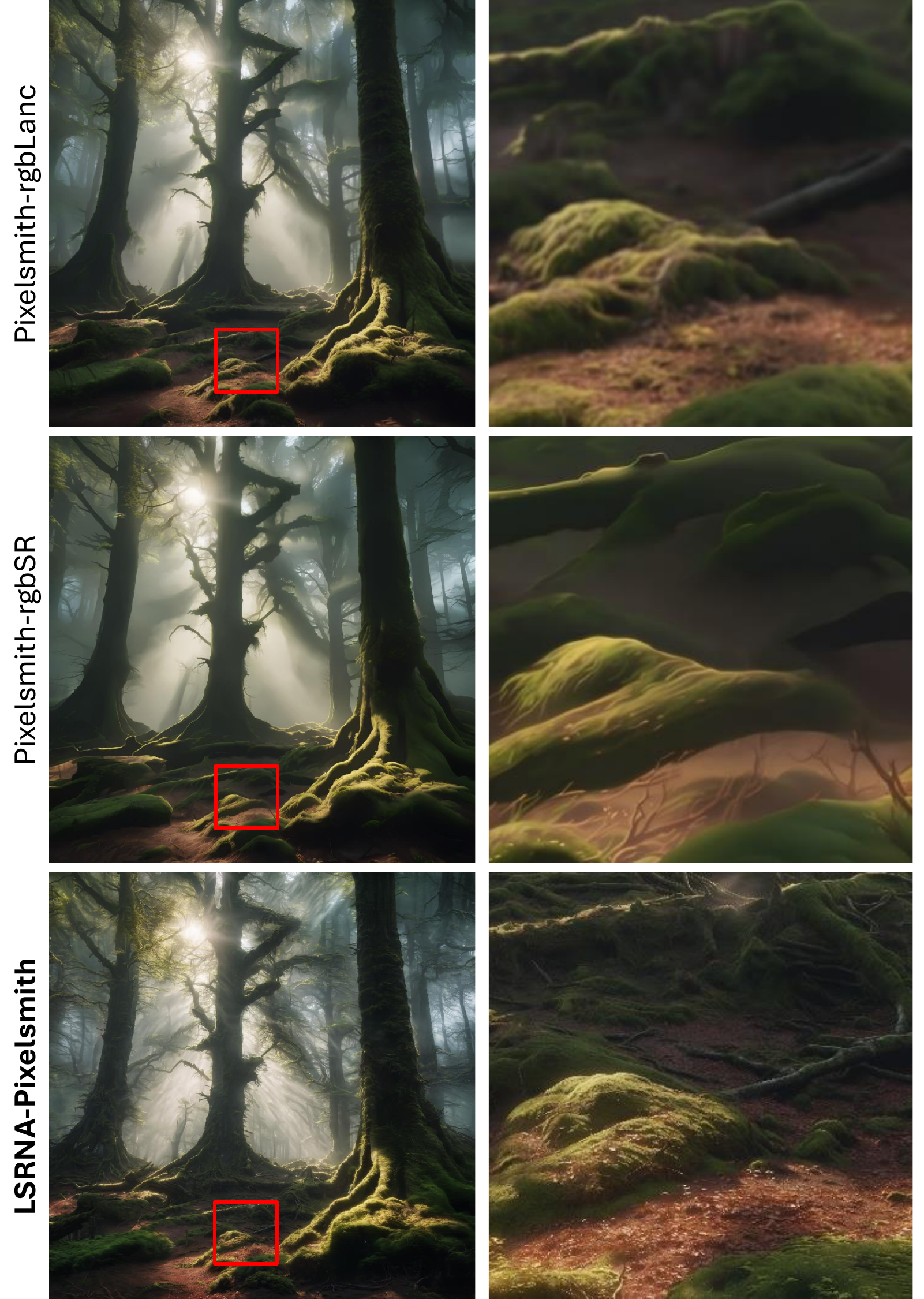}
    % \vspace{-2em}
    \caption{
    \textbf{RGB vs. Latent Space Upsampling for Pixelsmith on 64$\times$.}
    Prompt used is "A mysterious forest with tall, ancient trees and beams of sunlight filtering through the mist, detailed moss-covered roots, 8k".
    Best viewed \textbf{ZOOMED-IN}.
    }
    \label{fig:appendix_latent_importance4}
\end{figure*}
\clearpage
{
    \small
    \bibliographystyle{ieeenat_fullname}
    \bibliography{main}

\begin{thebibliography}{63}
\providecommand{\natexlab}[1]{#1}
\providecommand{\url}[1]{\texttt{#1}}
\expandafter\ifx\csname urlstyle\endcsname\relax
  \providecommand{\doi}[1]{doi: #1}\else
  \providecommand{\doi}{doi: \begingroup \urlstyle{rm}\Url}\fi

\bibitem[Bar-Tal et~al.(2023)Bar-Tal, Yariv, Lipman, and Dekel]{bar2023multidiffusion}
Omer Bar-Tal, Lior Yariv, Yaron Lipman, and Tali Dekel.
\newblock Multidiffusion: Fusing diffusion paths for controlled image generation.
\newblock \emph{arXiv preprint arXiv:2302.08113}, 2023.

\bibitem[Bi{\'n}kowski et~al.(2018)Bi{\'n}kowski, Sutherland, Arbel, and Gretton]{binkowski2018demystifying}
Miko{\l}aj Bi{\'n}kowski, Danica~J Sutherland, Michael Arbel, and Arthur Gretton.
\newblock Demystifying mmd gans.
\newblock \emph{arXiv preprint arXiv:1801.01401}, 2018.

\bibitem[Brooks et~al.(2023)Brooks, Holynski, and Efros]{brooks2023instructpix2pix}
Tim Brooks, Aleksander Holynski, and Alexei~A Efros.
\newblock Instructpix2pix: Learning to follow image editing instructions.
\newblock In \emph{Proceedings of the IEEE/CVF Conference on Computer Vision and Pattern Recognition}, pages 18392--18402, 2023.

\bibitem[Bruna et~al.(2015)Bruna, Sprechmann, and LeCun]{bruna2015super}
Joan Bruna, Pablo Sprechmann, and Yann LeCun.
\newblock Super-resolution with deep convolutional sufficient statistics.
\newblock \emph{arXiv preprint arXiv:1511.05666}, 2015.

\bibitem[Canny(1986)]{canny1986computational}
John Canny.
\newblock A computational approach to edge detection.
\newblock \emph{IEEE Transactions on Pattern Analysis and Machine Intelligence}, pages 679--698, 1986.

\bibitem[Chai et~al.(2022)Chai, Gharbi, Shechtman, Isola, and Zhang]{chai2022anyresolution}
Lucy Chai, Michael Gharbi, Eli Shechtman, Phillip Isola, and Richard Zhang.
\newblock Any-resolution training for high-resolution image synthesis.
\newblock In \emph{European Conference on Computer Vision}, 2022.

\bibitem[Chen et~al.(2024)Chen, Ge, Xie, Wu, Yao, Ren, Wang, Luo, Lu, and Li]{chen2024pixart}
Junsong Chen, Chongjian Ge, Enze Xie, Yue Wu, Lewei Yao, Xiaozhe Ren, Zhongdao Wang, Ping Luo, Huchuan Lu, and Zhenguo Li.
\newblock Pixart-sigma: Weak-to-strong training of diffusion transformer for 4k text-to-image generation.
\newblock \emph{arXiv preprint arXiv:2403.04692}, 2024.

\bibitem[Chen et~al.(2021)Chen, Liu, and Wang]{chen2021learning}
Yinbo Chen, Sifei Liu, and Xiaolong Wang.
\newblock Learning continuous image representation with local implicit image function.
\newblock In \emph{Proceedings of the IEEE/CVF Conference on Computer Vision and Pattern Recognition}, pages 8628--8638, 2021.

\bibitem[Dhariwal and Nichol(2021)]{dhariwal2021diffusion}
Prafulla Dhariwal and Alexander Nichol.
\newblock Diffusion models beat gans on image synthesis.
\newblock \emph{Advances in Neural Information Processing Systems}, 34:\penalty0 8780--8794, 2021.

\bibitem[Ding et~al.(2023)Ding, Zhang, Wu, and Tu]{ding2023patched}
Zheng Ding, Mengqi Zhang, Jiajun Wu, and Zhuowen Tu.
\newblock Patched denoising diffusion models for high-resolution image synthesis.
\newblock In \emph{International Conference on Learning Representations}, 2023.

\bibitem[Dosovitskiy(2020)]{dosovitskiy2020image}
Alexey Dosovitskiy.
\newblock An image is worth 16x16 words: Transformers for image recognition at scale.
\newblock \emph{arXiv preprint arXiv:2010.11929}, 2020.

\bibitem[Dosovitskiy and Brox(2016)]{dosovitskiy2016generating}
Alexey Dosovitskiy and Thomas Brox.
\newblock Generating images with perceptual similarity metrics based on deep networks.
\newblock \emph{Advances in Neural Information Processing Systems}, 29, 2016.

\bibitem[Du et~al.(2024)Du, Chang, Hospedales, Song, and Ma]{du2024demofusion}
Ruoyi Du, Dongliang Chang, Timothy Hospedales, Yi-Zhe Song, and Zhanyu Ma.
\newblock Demofusion: Democratising high-resolution image generation with no \$\$\$.
\newblock In \emph{Proceedings of the IEEE/CVF Conference on Computer Vision and Pattern Recognition}, pages 6159--6168, 2024.

\bibitem[Gabor(1946)]{gabor1946theory}
Dennis Gabor.
\newblock Theory of communication. part 1: The analysis of information.
\newblock \emph{Journal of the Institution of Electrical Engineers-part III: radio and communication engineering}, 93\penalty0 (26):\penalty0 429--441, 1946.

\bibitem[Gu et~al.(2023)Gu, Zhai, Zhang, Susskind, and Jaitly]{gu2023matryoshka}
Jiatao Gu, Shuangfei Zhai, Yizhe Zhang, Joshua~M Susskind, and Navdeep Jaitly.
\newblock Matryoshka diffusion models.
\newblock In \emph{International Conference on Learning Representations}, 2023.

\bibitem[Gu et~al.(2022)Gu, Chen, Bao, Wen, Zhang, Chen, Yuan, and Guo]{gu2022vector}
Shuyang Gu, Dong Chen, Jianmin Bao, Fang Wen, Bo Zhang, Dongdong Chen, Lu Yuan, and Baining Guo.
\newblock Vector quantized diffusion model for text-to-image synthesis.
\newblock In \emph{Proceedings of the IEEE/CVF Conference on Computer Vision and Pattern Recognition}, pages 10696--10706, 2022.

\bibitem[Guo et~al.(2024)Guo, He, Chen, Xia, Cun, Wang, Huang, Zhang, Wang, Chen, et~al.]{guo2024make}
Lanqing Guo, Yingqing He, Haoxin Chen, Menghan Xia, Xiaodong Cun, Yufei Wang, Siyu Huang, Yong Zhang, Xintao Wang, Qifeng Chen, et~al.
\newblock Make a cheap scaling: A self-cascade diffusion model for higher-resolution adaptation.
\newblock \emph{arXiv preprint arXiv:2402.10491}, 2024.

\bibitem[Haji-Ali et~al.(2024)Haji-Ali, Balakrishnan, and Ordonez]{haji2024elasticdiffusion}
Moayed Haji-Ali, Guha Balakrishnan, and Vicente Ordonez.
\newblock Elasticdiffusion: Training-free arbitrary size image generation through global-local content separation.
\newblock In \emph{Proceedings of the IEEE/CVF Conference on Computer Vision and Pattern Recognition}, pages 6603--6612, 2024.

\bibitem[He et~al.(2023)He, Yang, Chen, Cun, Xia, Zhang, Wang, He, Chen, and Shan]{he2023scalecrafter}
Yingqing He, Shaoshu Yang, Haoxin Chen, Xiaodong Cun, Menghan Xia, Yong Zhang, Xintao Wang, Ran He, Qifeng Chen, and Ying Shan.
\newblock Scalecrafter: Tuning-free higher-resolution visual generation with diffusion models.
\newblock In \emph{International Conference on Learning Representations}, 2023.

\bibitem[Hertz et~al.(2022)Hertz, Mokady, Tenenbaum, Aberman, Pritch, and Cohen-Or]{hertz2022prompt}
Amir Hertz, Ron Mokady, Jay Tenenbaum, Kfir Aberman, Yael Pritch, and Daniel Cohen-Or.
\newblock Prompt-to-prompt image editing with cross attention control.
\newblock \emph{arXiv preprint arXiv:2208.01626}, 2022.

\bibitem[Heusel et~al.(2017)Heusel, Ramsauer, Unterthiner, Nessler, and Hochreiter]{heusel2017gans}
Martin Heusel, Hubert Ramsauer, Thomas Unterthiner, Bernhard Nessler, and Sepp Hochreiter.
\newblock Gans trained by a two time-scale update rule converge to a local nash equilibrium.
\newblock \emph{Advances in Neural Information Processing Systems}, 30, 2017.

\bibitem[Ho et~al.(2020)Ho, Jain, and Abbeel]{ho2020denoising}
Jonathan Ho, Ajay Jain, and Pieter Abbeel.
\newblock Denoising diffusion probabilistic models.
\newblock \emph{Advances in Neural Information Processing Systems}, 33:\penalty0 6840--6851, 2020.

\bibitem[Ho et~al.(2022)Ho, Saharia, Chan, Fleet, Norouzi, and Salimans]{ho2022cascaded}
Jonathan Ho, Chitwan Saharia, William Chan, David~J Fleet, Mohammad Norouzi, and Tim Salimans.
\newblock Cascaded diffusion models for high fidelity image generation.
\newblock \emph{Journal of Machine Learning Research}, 23\penalty0 (47):\penalty0 1--33, 2022.

\bibitem[Hoogeboom et~al.(2023)Hoogeboom, Heek, and Salimans]{hoogeboom2023simple}
Emiel Hoogeboom, Jonathan Heek, and Tim Salimans.
\newblock simple diffusion: End-to-end diffusion for high resolution images.
\newblock In \emph{International Conference on Machine Learning}, pages 13213--13232. PMLR, 2023.

\bibitem[Huang et~al.(2024)Huang, Fang, Zhang, Song, Liu, Liu, and Li]{huang2024fouriscale}
Linjiang Huang, Rongyao Fang, Aiping Zhang, Guanglu Song, Si Liu, Yu Liu, and Hongsheng Li.
\newblock Fouriscale: A frequency perspective on training-free high-resolution image synthesis.
\newblock \emph{arXiv preprint arXiv:2403.12963}, 2024.

\bibitem[Johnson et~al.(2016)Johnson, Alahi, and Fei-Fei]{johnson2016perceptual}
Justin Johnson, Alexandre Alahi, and Li Fei-Fei.
\newblock Perceptual losses for real-time style transfer and super-resolution.
\newblock In \emph{European Conference on Computer Vision}, pages 694--711. Springer, 2016.

\bibitem[Kawar et~al.(2023)Kawar, Zada, Lang, Tov, Chang, Dekel, Mosseri, and Irani]{kawar2023imagic}
Bahjat Kawar, Shiran Zada, Oran Lang, Omer Tov, Huiwen Chang, Tali Dekel, Inbar Mosseri, and Michal Irani.
\newblock Imagic: Text-based real image editing with diffusion models.
\newblock In \emph{Proceedings of the IEEE/CVF Conference on Computer Vision and Pattern Recognition}, pages 6007--6017, 2023.

\bibitem[Kingma(2014)]{kingma2014adam}
Diederik~P Kingma.
\newblock Adam: A method for stochastic optimization.
\newblock \emph{arXiv preprint arXiv:1412.6980}, 2014.

\bibitem[Kuznetsova et~al.(2020)Kuznetsova, Rom, Alldrin, Uijlings, Krasin, Pont-Tuset, Kamali, Popov, Malloci, Kolesnikov, et~al.]{kuznetsova2020open}
Alina Kuznetsova, Hassan Rom, Neil Alldrin, Jasper Uijlings, Ivan Krasin, Jordi Pont-Tuset, Shahab Kamali, Stefan Popov, Matteo Malloci, Alexander Kolesnikov, et~al.
\newblock The open images dataset v4: Unified image classification, object detection, and visual relationship detection at scale.
\newblock \emph{International Journal of Computer Vision}, 128\penalty0 (7):\penalty0 1956--1981, 2020.

\bibitem[Lanczos(1964)]{lanczos1964evaluation}
Cornelius Lanczos.
\newblock Evaluation of noisy data.
\newblock \emph{Journal of the Society for Industrial and Applied Mathematics, Series B: Numerical Analysis}, 1\penalty0 (1):\penalty0 76--85, 1964.

\bibitem[Lefaudeux et~al.(2022)Lefaudeux, Massa, Liskovich, Xiong, Caggiano, Naren, Xu, Hu, Tintore, Zhang, et~al.]{lefaudeux2022xformers}
Benjamin Lefaudeux, Francisco Massa, Diana Liskovich, Wenhan Xiong, Vittorio Caggiano, Sean Naren, Min Xu, Jieru Hu, Marta Tintore, Susan Zhang, et~al.
\newblock xformers: A modular and hackable transformer modelling library, 2022.

\bibitem[Li et~al.(2023)Li, Li, Savarese, and Hoi]{li2023blip}
Junnan Li, Dongxu Li, Silvio Savarese, and Steven Hoi.
\newblock Blip-2: Bootstrapping language-image pre-training with frozen image encoders and large language models.
\newblock In \emph{International Conference on Machine Learning}, pages 19730--19742. PMLR, 2023.

\bibitem[Liang et~al.(2021)Liang, Cao, Sun, Zhang, Van~Gool, and Timofte]{liang2021swinir}
Jingyun Liang, Jiezhang Cao, Guolei Sun, Kai Zhang, Luc Van~Gool, and Radu Timofte.
\newblock Swinir: Image restoration using swin transformer.
\newblock In \emph{Proceedings of the IEEE/CVF International Conference on Computer Vision}, pages 1833--1844, 2021.

\bibitem[Lin et~al.(2024)Lin, Lin, Zhao, and Ji]{lin2024accdiffusion}
Zhihang Lin, Mingbao Lin, Meng Zhao, and Rongrong Ji.
\newblock Accdiffusion: An accurate method for higher-resolution image generation.
\newblock \emph{arXiv preprint arXiv:2407.10738}, 2024.

\bibitem[Marr and Hildreth(1980)]{marr1980theory}
David Marr and Ellen Hildreth.
\newblock Theory of edge detection.
\newblock \emph{Proceedings of the Royal Society of London. Series B. Biological Sciences}, 207\penalty0 (1167):\penalty0 187--217, 1980.

\bibitem[Meng et~al.(2021)Meng, He, Song, Song, Wu, Zhu, and Ermon]{meng2021sdedit}
Chenlin Meng, Yutong He, Yang Song, Jiaming Song, Jiajun Wu, Jun-Yan Zhu, and Stefano Ermon.
\newblock Sdedit: Guided image synthesis and editing with stochastic differential equations.
\newblock \emph{arXiv preprint arXiv:2108.01073}, 2021.

\bibitem[Mokady et~al.(2023)Mokady, Hertz, Aberman, Pritch, and Cohen-Or]{mokady2023null}
Ron Mokady, Amir Hertz, Kfir Aberman, Yael Pritch, and Daniel Cohen-Or.
\newblock Null-text inversion for editing real images using guided diffusion models.
\newblock In \emph{Proceedings of the IEEE/CVF Conference on Computer Vision and Pattern Recognition}, pages 6038--6047, 2023.

\bibitem[Nichol et~al.(2021)Nichol, Dhariwal, Ramesh, Shyam, Mishkin, McGrew, Sutskever, and Chen]{nichol2021glide}
Alex Nichol, Prafulla Dhariwal, Aditya Ramesh, Pranav Shyam, Pamela Mishkin, Bob McGrew, Ilya Sutskever, and Mark Chen.
\newblock Glide: Towards photorealistic image generation and editing with text-guided diffusion models.
\newblock \emph{arXiv preprint arXiv:2112.10741}, 2021.

\bibitem[Podell et~al.(2023)Podell, English, Lacey, Blattmann, Dockhorn, M{\"u}ller, Penna, and Rombach]{podell2023sdxl}
Dustin Podell, Zion English, Kyle Lacey, Andreas Blattmann, Tim Dockhorn, Jonas M{\"u}ller, Joe Penna, and Robin Rombach.
\newblock Sdxl: Improving latent diffusion models for high-resolution image synthesis.
\newblock \emph{arXiv preprint arXiv:2307.01952}, 2023.

\bibitem[Radford et~al.(2021)Radford, Kim, Hallacy, Ramesh, Goh, Agarwal, Sastry, Askell, Mishkin, Clark, et~al.]{radford2021learning}
Alec Radford, Jong~Wook Kim, Chris Hallacy, Aditya Ramesh, Gabriel Goh, Sandhini Agarwal, Girish Sastry, Amanda Askell, Pamela Mishkin, Jack Clark, et~al.
\newblock Learning transferable visual models from natural language supervision.
\newblock In \emph{International Conference on Machine Learning}, pages 8748--8763. PMLR, 2021.

\bibitem[Ramesh et~al.(2022)Ramesh, Dhariwal, Nichol, Chu, and Chen]{ramesh2022hierarchical}
Aditya Ramesh, Prafulla Dhariwal, Alex Nichol, Casey Chu, and Mark Chen.
\newblock Hierarchical text-conditional image generation with clip latents.
\newblock \emph{arXiv preprint arXiv:2204.06125}, 1\penalty0 (2):\penalty0 3, 2022.

\bibitem[Ren et~al.(2024)Ren, Li, Chen, Pei, Shao, Guo, Peng, Song, and Zhu]{ren2024ultrapixel}
Jingjing Ren, Wenbo Li, Haoyu Chen, Renjing Pei, Bin Shao, Yong Guo, Long Peng, Fenglong Song, and Lei Zhu.
\newblock Ultrapixel: Advancing ultra-high-resolution image synthesis to new peaks.
\newblock \emph{arXiv preprint arXiv:2407.02158}, 2024.

\bibitem[Rombach et~al.(2022)Rombach, Blattmann, Lorenz, Esser, and Ommer]{rombach2022high}
Robin Rombach, Andreas Blattmann, Dominik Lorenz, Patrick Esser, and Bj{\"o}rn Ommer.
\newblock High-resolution image synthesis with latent diffusion models.
\newblock In \emph{Proceedings of the IEEE/CVF Conference on Computer Vision and Pattern Recognition}, pages 10684--10695, 2022.

\bibitem[Ruiz et~al.(2023)Ruiz, Li, Jampani, Pritch, Rubinstein, and Aberman]{ruiz2023dreambooth}
Nataniel Ruiz, Yuanzhen Li, Varun Jampani, Yael Pritch, Michael Rubinstein, and Kfir Aberman.
\newblock Dreambooth: Fine tuning text-to-image diffusion models for subject-driven generation.
\newblock In \emph{Proceedings of the IEEE/CVF Conference on Computer Vision and Pattern Recognition}, pages 22500--22510, 2023.

\bibitem[Saharia et~al.(2022{\natexlab{a}})Saharia, Chan, Chang, Lee, Ho, Salimans, Fleet, and Norouzi]{saharia2022palette}
Chitwan Saharia, William Chan, Huiwen Chang, Chris Lee, Jonathan Ho, Tim Salimans, David Fleet, and Mohammad Norouzi.
\newblock Palette: Image-to-image diffusion models.
\newblock In \emph{ACM SIGGRAPH 2022 conference proceedings}, pages 1--10, 2022{\natexlab{a}}.

\bibitem[Saharia et~al.(2022{\natexlab{b}})Saharia, Chan, Saxena, Li, Whang, Denton, Ghasemipour, Gontijo~Lopes, Karagol~Ayan, Salimans, et~al.]{saharia2022photorealistic}
Chitwan Saharia, William Chan, Saurabh Saxena, Lala Li, Jay Whang, Emily~L Denton, Kamyar Ghasemipour, Raphael Gontijo~Lopes, Burcu Karagol~Ayan, Tim Salimans, et~al.
\newblock Photorealistic text-to-image diffusion models with deep language understanding.
\newblock \emph{Advances in Neural Information Processing Systems}, 35:\penalty0 36479--36494, 2022{\natexlab{b}}.

\bibitem[Saharia et~al.(2022{\natexlab{c}})Saharia, Ho, Chan, Salimans, Fleet, and Norouzi]{saharia2022image}
Chitwan Saharia, Jonathan Ho, William Chan, Tim Salimans, David~J Fleet, and Mohammad Norouzi.
\newblock Image super-resolution via iterative refinement.
\newblock \emph{IEEE Transactions on Pattern Analysis and Machine Intelligence}, 45\penalty0 (4):\penalty0 4713--4726, 2022{\natexlab{c}}.

\bibitem[Salimans et~al.(2016)Salimans, Goodfellow, Zaremba, Cheung, Radford, and Chen]{salimans2016improved}
Tim Salimans, Ian Goodfellow, Wojciech Zaremba, Vicki Cheung, Alec Radford, and Xi Chen.
\newblock Improved techniques for training gans.
\newblock \emph{Advances in Neural Information Processing Systems}, 29, 2016.

\bibitem[Scharr(2000)]{scharr2000optimal}
Hanno Scharr.
\newblock Optimal operators in digital image processing.
\newblock 2000.

\bibitem[Shi et~al.(2024)Shi, Li, Zhang, He, Gong, and Zheng]{shi2024resmaster}
Shuwei Shi, Wenbo Li, Yuechen Zhang, Jingwen He, Biao Gong, and Yinqiang Zheng.
\newblock Resmaster: Mastering high-resolution image generation via structural and fine-grained guidance.
\newblock \emph{arXiv preprint arXiv:2406.16476}, 2024.

\bibitem[Shi et~al.(2016)Shi, Caballero, Husz{\'a}r, Totz, Aitken, Bishop, Rueckert, and Wang]{shi2016real}
Wenzhe Shi, Jose Caballero, Ferenc Husz{\'a}r, Johannes Totz, Andrew~P Aitken, Rob Bishop, Daniel Rueckert, and Zehan Wang.
\newblock Real-time single image and video super-resolution using an efficient sub-pixel convolutional neural network.
\newblock In \emph{Proceedings of the IEEE conference on computer vision and pattern recognition}, pages 1874--1883, 2016.

\bibitem[Si et~al.(2024)Si, Huang, Jiang, and Liu]{si2024freeu}
Chenyang Si, Ziqi Huang, Yuming Jiang, and Ziwei Liu.
\newblock Freeu: Free lunch in diffusion u-net.
\newblock In \emph{Proceedings of the IEEE/CVF Conference on Computer Vision and Pattern Recognition}, pages 4733--4743, 2024.

\bibitem[Song et~al.(2020)Song, Meng, and Ermon]{song2020denoising}
Jiaming Song, Chenlin Meng, and Stefano Ermon.
\newblock Denoising diffusion implicit models.
\newblock \emph{arXiv preprint arXiv:2010.02502}, 2020.

\bibitem[Szegedy et~al.(2015)Szegedy, Liu, Jia, Sermanet, Reed, Anguelov, Erhan, Vanhoucke, and Rabinovich]{szegedy2015going}
Christian Szegedy, Wei Liu, Yangqing Jia, Pierre Sermanet, Scott Reed, Dragomir Anguelov, Dumitru Erhan, Vincent Vanhoucke, and Andrew Rabinovich.
\newblock Going deeper with convolutions.
\newblock In \emph{Proceedings of the IEEE/CVF Conference on Computer Vision and Pattern Recognition}, pages 1--9, 2015.

\bibitem[Tragakis et~al.(2024)Tragakis, Aversa, Kaul, Murray-Smith, and Faccio]{tragakis2024one}
Athanasios Tragakis, Marco Aversa, Chaitanya Kaul, Roderick Murray-Smith, and Daniele Faccio.
\newblock Is one gpu enough? pushing image generation at higher-resolutions with foundation models.
\newblock \emph{arXiv preprint arXiv:2406.07251}, 2024.

\bibitem[Wang et~al.(2021)Wang, Xie, Dong, and Shan]{wang2021real}
Xintao Wang, Liangbin Xie, Chao Dong, and Ying Shan.
\newblock Real-esrgan: Training real-world blind super-resolution with pure synthetic data.
\newblock In \emph{Proceedings of the IEEE/CVF International Conference on Computer Vision}, pages 1905--1914, 2021.

\bibitem[Xie et~al.(2024)Xie, Chen, Chen, Cai, Lin, Zhang, Li, Lu, and Han]{xie2024sana}
Enze Xie, Junsong Chen, Junyu Chen, Han Cai, Yujun Lin, Zhekai Zhang, Muyang Li, Yao Lu, and Song Han.
\newblock Sana: Efficient high-resolution image synthesis with linear diffusion transformers.
\newblock \emph{arXiv preprint arXiv:2410.10629}, 2024.

\bibitem[Zhang et~al.(2021)Zhang, Liang, Van~Gool, and Timofte]{zhang2021designing}
Kai Zhang, Jingyun Liang, Luc Van~Gool, and Radu Timofte.
\newblock Designing a practical degradation model for deep blind image super-resolution.
\newblock In \emph{Proceedings of the IEEE/CVF International Conference on Computer Vision}, pages 4791--4800, 2021.

\bibitem[Zhang et~al.(2023{\natexlab{a}})Zhang, Rao, and Agrawala]{zhang2023adding}
Lvmin Zhang, Anyi Rao, and Maneesh Agrawala.
\newblock Adding conditional control to text-to-image diffusion models.
\newblock In \emph{Proceedings of the IEEE/CVF International Conference on Computer Vision}, pages 3836--3847, 2023{\natexlab{a}}.

\bibitem[Zhang et~al.(2018{\natexlab{a}})Zhang, Isola, Efros, Shechtman, and Wang]{zhang2018unreasonable}
Richard Zhang, Phillip Isola, Alexei~A Efros, Eli Shechtman, and Oliver Wang.
\newblock The unreasonable effectiveness of deep features as a perceptual metric.
\newblock In \emph{Proceedings of the IEEE/CVF Conference on Computer Vision and Pattern Recognition}, pages 586--595, 2018{\natexlab{a}}.

\bibitem[Zhang et~al.(2023{\natexlab{b}})Zhang, Chen, Zhao, Chen, Tang, Chen, Cao, and Liang]{zhang2023hidiffusion}
Shen Zhang, Zhaowei Chen, Zhenyu Zhao, Zhenyuan Chen, Yao Tang, Yuhao Chen, Wengang Cao, and Jiajun Liang.
\newblock Hidiffusion: Unlocking high-resolution creativity and efficiency in low-resolution trained diffusion models.
\newblock \emph{arXiv preprint arXiv:2311.17528}, 2023{\natexlab{b}}.

\bibitem[Zhang et~al.(2018{\natexlab{b}})Zhang, Li, Li, Wang, Zhong, and Fu]{zhang2018image}
Yulun Zhang, Kunpeng Li, Kai Li, Lichen Wang, Bineng Zhong, and Yun Fu.
\newblock Image super-resolution using very deep residual channel attention networks.
\newblock In \emph{European Conference on Computer Vision}, pages 286--301, 2018{\natexlab{b}}.

\bibitem[Zheng et~al.(2024)Zheng, Guo, Deng, Han, Li, Xu, and Xu]{zheng2024any}
Qingping Zheng, Yuanfan Guo, Jiankang Deng, Jianhua Han, Ying Li, Songcen Xu, and Hang Xu.
\newblock Any-size-diffusion: Toward efficient text-driven synthesis for any-size hd images.
\newblock In \emph{Proceedings of the AAAI Conference on Artificial Intelligence}, pages 7571--7578, 2024.

\end{thebibliography}
}
\end{document}